\documentclass[lettersize,journal]{IEEEtran}
\usepackage{amsmath,amsfonts}
\usepackage{algorithmic}
\usepackage{algorithm}
\usepackage{booktabs}
\usepackage{multirow}
\usepackage{array}
\usepackage{textcomp}
\usepackage{stfloats}
\usepackage{verbatim}
\usepackage{graphicx}
\usepackage{makecell}
\PassOptionsToPackage{hyphens}{url}
\usepackage{xurl}
\usepackage{pifont}
\usepackage{tabularx}
\usepackage[colorlinks=true, citecolor=blue, linkcolor=blue, urlcolor=blue, breaklinks=true]{hyperref}
\newcommand{\transpose}{\mathsf{T}}

\begin{document}

\title{Inland-LOAM: Voxel-Based Structural 
Semantic LiDAR Odometry and Mapping for Inland Waterway Navigation}

\author{Zhongbi Luo, Yunjia Wang, Jan Swevers, Peter Slaets and Herman Bruyninckx

\thanks{\copyright~2026 IEEE. Personal use of this material is permitted. Permission from IEEE must be obtained for all other uses, in any current or future media, including reprinting/republishing this material for advertising or promotional purposes, creating new collective works, for resale or redistribution to servers or lists, or reuse of any copyrighted component of this work in other works.}
\thanks{This work has received funding from the European Union's Horizon 2020 research and innovation programme under the Marie Sk\l{}odowska-Curie grant agreement No.~955768. Project website: \url{https://etn-autobarge.eu/}. (Corresponding author: Zhongbi Luo.)}
\thanks{Zhongbi Luo, Peter Slaets, Jan Swevers and Herman Bruyninckx are with the Division of Robotics, Automation and Mechatronics in the Department of Mechanical Engineering, KU Leuven, 3001 Leuven, Belgium (e-mail: zhongbi.luo@kuleuven.be; peter.slaets@kuleuven.be; jan.swevers@kuleuven.be; herman.bruyninckx@kuleuven.be).}
\thanks{Yunjia Wang is with the Division of Declarative Languages and Artificial Intelligence (DTAI) in the Department of Computer Science, KU Leuven, 8200 Bruges, Belgium (e-mail: yunjia.wang@kuleuven.be).}
\thanks{Yunjia Wang, Jan Swevers and Herman Bruyninckx are also with Flanders Make@KU Leuven, 3001 Leuven, Belgium. Herman Bruyninckx is also with the Department of Mechanical Engineering, Eindhoven University of Technology, 5612 AZ Eindhoven, The Netherlands.}
\thanks{Digital Object Identifier 10.1109/TITS.2026.3756570}}

\markboth{IEEE TRANSACTIONS ON INTELLIGENT TRANSPORTATION SYSTEMS}%
{Luo \MakeLowercase{\textit{et al.}}: Inland-LOAM for Inland Waterway Navigation}

\maketitle

\begin{abstract}
Accurate and up-to-date geospatial information is crucial for enhancing the safety and autonomy of Inland Waterway Transport. Existing Inland Electronic Navigational Charts often lack real-time detail, and conventional LiDAR-based Simultaneous Localization and Mapping methods struggle with waterway characteristics like unstructured geometric features and water surface reflections. These challenges lead to significant localization drift and produce point cloud maps lacking the semantic richness required for autonomous decision-making. This paper introduces a comprehensive LiDAR odometry
and mapping framework for inland waterway navigation (Inland-LOAM). We present an improved feature extraction method adapted to unique waterway geometries, combined with a joint optimization that incorporates the water surface as a global planar constraint to mitigate drift. We also propose an innovative pipeline that transforms dense 3D point cloud outputs into 2D structural semantic maps. By constructing semantic voxel grids and performing geometric analyses (roughness, planarity, and slope), our system classifies the environment into meaningful structural classes and supports real-time computation of navigation-relevant quantities such as vertical bridge clearances. An automated module then efficiently extracts shoreline boundaries and exports them as lightweight geospatial data that can be integrated with standard navigational charts. Extensive evaluations on a diverse, real-world dataset demonstrate that Inland-LOAM achieves higher localization accuracy than the evaluated baselines. The generated maps and shorelines align with real-world conditions, providing useful information to enhance navigational situational awareness.

\end{abstract}



\section{Introduction}
\label{sec1}

Inland Waterway Transport (IWT) constitutes an essential component of Europe's freight infrastructure, spanning a network exceeding 41,000 km, interlinking major cities and industrial hubs across 13 interconnected Member States \cite{EC_inland_waterways}. As efforts increase to shift freight from congested road and rail networks, the importance of accurate geospatial information and detailed environmental models for managing and navigating these waterways grows \cite{EC_RIS_Directive_2024}. 
However, navigating these waterways involves safety risks, with studies identifying human error as the primary cause in up to 70\% of maritime incidents \cite{galierikova2019human}. In this context, advancements in automation, especially the deployment of uncrewed surface vehicles (USVs), are driving a transformation in inland waterway transport. USVs play a crucial role not only by mitigating the risks associated with human factors, but also by promising improved cargo transport efficiency and facilitating innovative logistics services, including operations within challenging environments such as congested urban canal networks \cite{ROBOT_I}. However, the effective operation of USVs is challenging due to the complex and dynamic nature of inland waterways. These are confined channels with locks and (movable and fixed) bridges and limited water depth, which often create navigation bottlenecks \cite{segovia2025dynamic}. Safe and efficient navigation mainly depends on robust systems for perception, localization, and mapping. These systems must enable the vessel to reliably recognize diverse surroundings, including shorelines, infrastructure, vegetation, and overhead obstacles such as bridges, to operate safely.

Traditionally, navigation on inland waterways relies heavily on Global Navigation Satellite Systems (GNSS) for positioning, complemented by standard nautical charts like Inland Electronic Navigational Charts (IENCs) for contextual spatial data \cite{kamal2017inland}. However, GNSS frequently suffers from severe attenuation or complete outages in typical IWT scenarios, particularly when vessels pass under bridges, navigate narrow urban canals flanked by tall buildings (``urban canyons''), or travel along waterways with dense vegetation canopies \cite{wang2020Roboat_II, kriechbaumer2015quantitative}. This documented unreliability highlights the need for onboard localization systems that work independently of GNSS \cite{fuentes2015visual}. Furthermore, existing geospatial data products, such as IENCs, while providing important routing and regulatory data, often lack the spatial detail or real-time updates required for accurate autonomous maneuvering and perception. Representations of shorelines and their surrounding features, including vegetation, shoreline constructions, and other obstacles, are often generalized or outdated in existing maps \cite{senne9606960, pieper2024conceptual}. A key limitation of IENCs is the representation of safety parameters, such as vertical bridge clearance, using static data. This approach fails to account for dynamic water level variations, a significant operational challenge in IWT affecting actual clearances \cite{hosch2023high}. As a result, relying on static, low-resolution geospatial information hinders the deployment of advanced autonomous functions.

LiDAR-based Simultaneous Localization and Mapping (SLAM) has developed as a strong technique in terrestrial robotics and autonomous driving,
making it possible to build precise 3D maps and achieve accurate localization even in environments without GNSS \cite{Indoor_slam_zhao2021general, GNSS_DENIED_outdoor_9199302}.  Applying LiDAR SLAM to the IWT is a viable way to overcome the limitations of previous approaches. However, traditional LiDAR SLAM algorithms face particular challenges in this context. The main challenges come from the structure of the environment: unlike urban scenes with abundant planar facades (buildings) and consistent features (lane markings), inland waterways predominantly feature irregular, natural shorelines often covered with vegetation \cite{hosch2023high}. This lack of structured geometry and standardized visual cues affects the reliable extraction of edges and planar features required by algorithms like LOAM \cite{zhang2014loam} and LeGO-LOAM \cite{shan2018lego}. Although this lack of structured geometry presents a challenge, these environments simultaneously offer a unique and powerful geometric prior that is often absent in terrestrial scenarios: the water surface itself. Rather than being treated as a source of interference, we contend that when sufficient LiDAR returns are available, the inland water surface can provide a broad and continuous geometric reference, mainly for roll, pitch, and height consistency. In addition to these algorithmic challenges, the development and evaluation of SLAM algorithms for IWT are hindered by the lack of suitable benchmark datasets.

Beyond achieving accurate geometric reconstruction, effective scene understanding demands semantic information, e.g., distinguishing navigable water from vegetated banks, identifying a potential docking quay wall, or determining safe passage under a bridge. Standard SLAM methods typically do not provide this layer of structural semantic information. Finally, a gap exists in translating the rich, high-resolution data from SLAM systems into formats aligned with navigation standards.

\begin{figure}[!t]
  \centering
  \includegraphics[width=0.9\columnwidth]{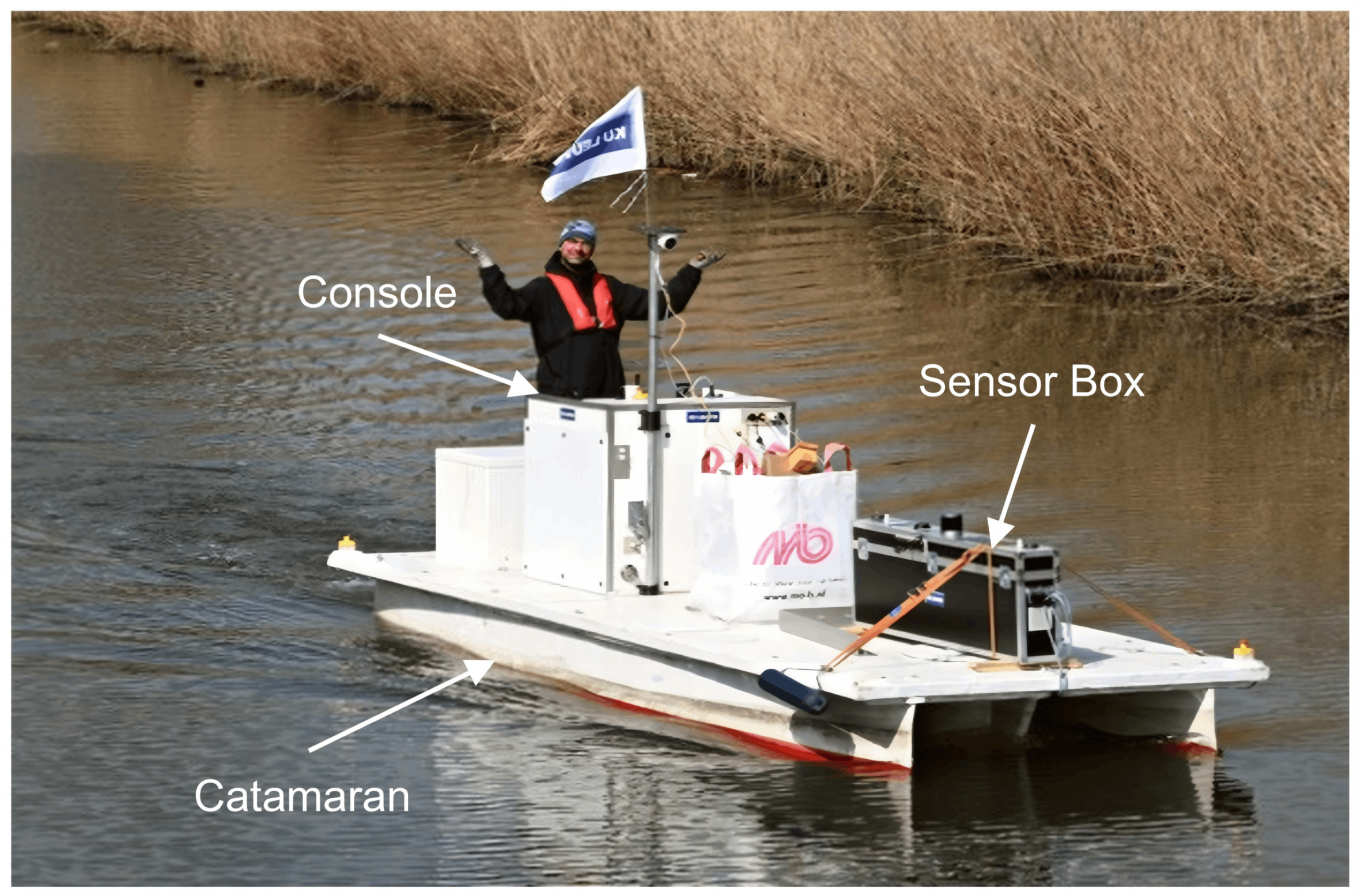}
  \caption{Our experimental platform, a catamaran named the Maverick, was equipped with a Sensor Box for situational awareness and data collection. Image source: \cite{Wouters2023}.}
  \label{fig:Maverick}
\end{figure}

To address these challenges, we propose Inland-LOAM, a novel voxel-based system designed for structural semantic mapping in inland waterway environments. This method achieves robust LiDAR-based localization and mapping adapted for water surfaces, extracts structural semantic information using geometric analysis within a voxel map, and generates enhanced environmental representations suitable for navigation and integration with standard chart formats. The contributions of this paper are as follows:
\begin{enumerate}
    \item Inland-LOAM is specifically enhanced for robustness in GNSS-denied inland waterway environments through improved feature extraction performance and the incorporation of water surface optimization to mitigate the prevalent vertical drift in inland waterway navigation.
    \item A novel voxel-based structural semantic mapping approach using roughness, planarity, and slope analysis to generate 2D maps and calculate real-time bridge clearance.
    \item An automated map conversion method that extracts shorelines and bridge regions from the structural semantic map and exports lightweight geospatial data for chart update systems.
    \item  The proposed framework was demonstrated on a diverse set of real-world inland waterway scenes, including ports, urban canals, vegetated banks, bridges, and lifting gates. The experimental platform used for data collection is shown in Figure~\ref{fig:Maverick}.  The source code and data download links are publicly available at \url{https://github.com/ZhongbiLuo/Inland-LOAM} to support further research.

\end{enumerate}

\section{Related Work}
\label{sec2}

This section reviews existing literature relevant to LiDAR-based mapping and semantic understanding in the context of inland waterways, focusing on inland mapping methodologies, LiDAR odometry techniques, semantic scene understanding, and map integration with navigation standards.

%% Use \subsection commands to start a subsection.
\subsection{Mapping in Inland Waterway Environments}
\label{subsec1}

LiDAR-based onboard mapping offers a promising solution for detailed mapping in inland scenarios \cite{luo2024enhancing, dietmayer2023smart, wang2025dr}. However, deploying LiDAR SLAM in IWT presents distinct complexities compared to established terrestrial applications \cite{bavle2022situational, zou2021comparative}:

Water Surface Interference: Reflections on the surface can distort LiDAR returns, causing noisy or inaccurate measurements. Also, signal absorption along the Z-axis weakens vertical constraints \cite{chen2026heterogeneous}. Due to continuous motion induced by waves and currents, vessels cannot be modeled using non-holonomic motion constraints \cite{10773240}.

Long Trajectories and Sparse Loops: The nature of waterway transport involves long traverses with infrequent returns to previously mapped areas, demanding high odometry accuracy due to limited opportunities for loop closure \cite{10773240}.

Data Scarcity: The robust evaluation of SLAM algorithms is hampered by the lack of comprehensive public datasets that capture the various conditions (narrow canals, locks, vegetation, bridges) and operational realities of European IWT; existing datasets like USVInland \cite{cheng2021we} often have limitations in scope or relevance.

\subsection{LiDAR Odometry and Mapping Variants}
\label{subsec2}

LOAM \cite{zhang2014loam} pioneered efficient, real-time LiDAR odometry based on the roughness of each point on its respective ring.
LeGO-LOAM \cite{shan2018lego} built upon LOAM by incorporating point cloud segmentation to separate ground points from other objects and clustering non-ground points. This allows feature extraction to focus on more salient objects and significantly reduces computational cost, making the method suitable for platforms with limited processing power. Subsequent works, such as F-LOAM \cite{wang2021f}, further focused on developing lightweight LOAM variants with optimized efficiency. For flat-ground scenarios, HDL-SLAM \cite{koide2019portable} employed graph optimization using NDT scan matching integrated with ground plane constraints. Ground-SLAM \cite{wei2021ground} specifically addressed structured multi-floor environments by extracting ground segments and modeling them as planar landmarks via Closest Point (CP) parameterization. Recently, KISS-ICP \cite{vizzo2023kiss} achieved high performance through simple but effective point-to-point registration and accurate motion compensation. However, methods explicitly incorporating ground plane constraints typically assume the existence of a continuous solid ground. Additionally, ICP-based approaches, despite their computational efficiency, often overlook the richer geometric information provided by feature-based methods.

\subsection{Semantic SLAM and Scene Understanding}
\label{subsec3}

 Semantic SLAM commonly relies on Deep Learning (DL).  SuMa++ \cite{chen2019suma++} used Fully Convolutional Networks (FCNs) \cite{milioto2019rangenet++} for segmentation that allows filtering of dynamic objects. Similarly, Gong et al. \cite{gong2019mapping} integrated LiDAR SLAM with a PointNet-based segmentation head to obtain semantic modeling in GNSS-denied underground parking lots. Another study \cite{wei2024semantic} leveraged a transformer-based model for integrated feature extraction and segmentation, enabling selective processing focused on static scene components to enhance localization accuracy. More generally, LiDAR semantic segmentation methods \cite{hu2020randla,puy2023using,xu2025frnet} provide point-level scene labels that can support semantic mapping and serve as learning-based baselines. While effective for object recognition, such DL methods typically rely on large labeled datasets.
 
In contrast, geometry-based approaches derive semantic understanding without extensive annotations. $\mathcal {PLC}$-LiSLAM \cite{zhou2022mathcal} detected geometric primitives, such as planes, lines, and cylinders, as stable landmarks for SLAM optimization. Similarly, the method in \cite{de2023air} achieved forest localization by clustering to identify trees and matching them to aerial canopy maps. These methods leverage inherent geometric structures for environmental understanding, avoiding the need for learned object appearance models.

\subsection{Inland Navigation Standards and Map Integration}
\label{subsec4}

IENCs are vital for navigation on European inland waterways, delivering essential details on waterway layouts, aids to navigation, locks, and bridges \cite{pfliegl2006increasing}. Recent studies have explored enhancing IENCs with sensor data. For example, Van Baelen et al. \cite{senne9606960} used onboard LiDAR to enhance situational awareness for inland vessels, detecting environmental changes using a k-d tree and aligning them with IENC data. Pieper and Hahn \cite{pieper2024conceptual} proposed a 3D LiDAR sensor fusion method to improve ENC accuracy in harbor areas. In our prior work, Luo et al. \cite{luo2024enhancing} dynamically updated IENCs using clustering techniques on LiDAR data. While these efforts highlight the potential of LiDAR for IENC updates, they often depend on direct projection and ad-hoc feature fitting. Similarly, Fredriksson et al. \cite{fredriksson2024voxel} converted voxel maps into 2D occupancy grids for path planning by projecting verified free space and preserving height and slope metadata, suggesting a promising pathway for efficient 3D-to-2D map conversion. However, existing research either validates isolated ENC objects or compresses 2D maps for robotic planning, lacking an end-to-end algorithm to transform LiDAR-derived structural and semantic data into standardized geospatial overlays suitable for IENCs.

\section{Methodology}
\label{sec3}

\subsection{System Overview}
\label{subsec3_1}

\begin{figure*}[!t]
  \centering
  \includegraphics[width=0.95\textwidth]{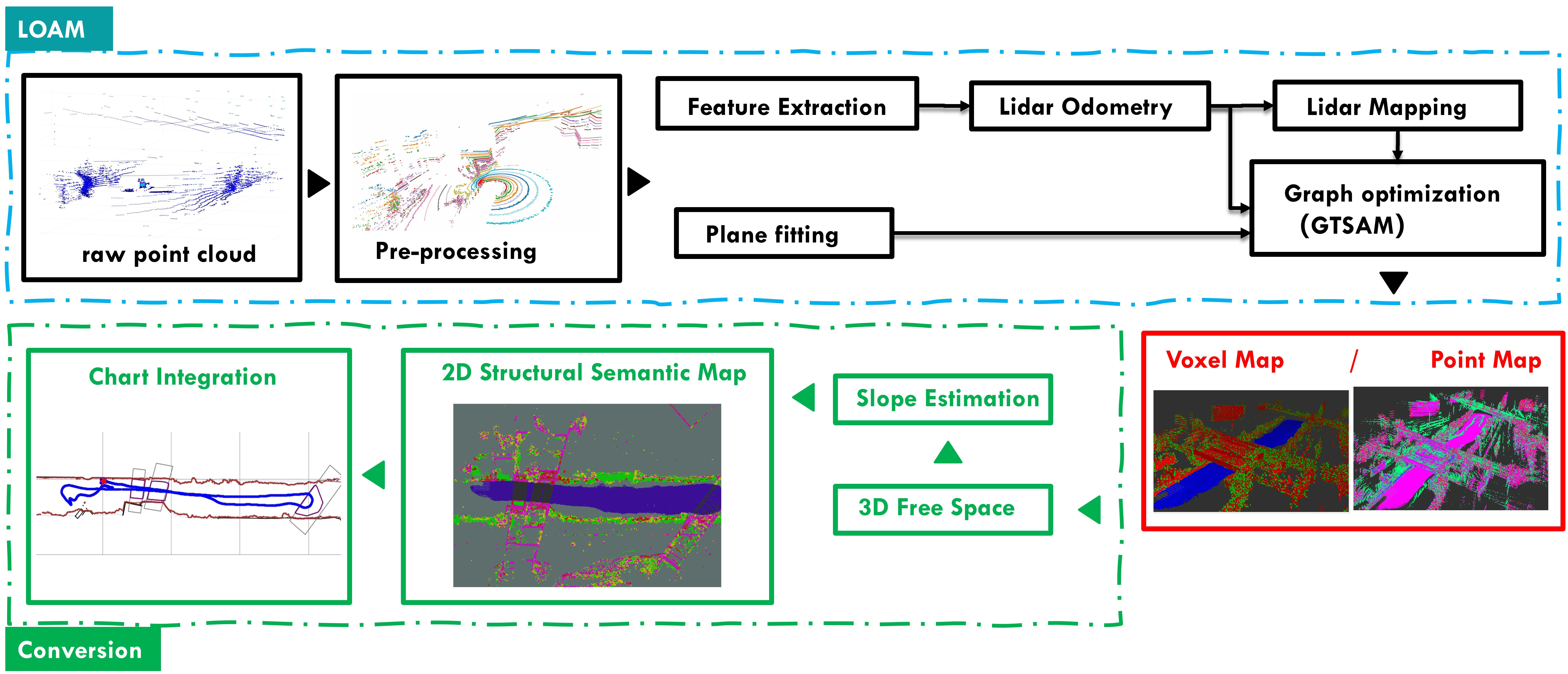}
  \caption{Overview of the Inland-LOAM framework, which processes LiDAR data to generate structural semantic maps and chart integration outputs for inland waterways. The LOAM pipeline (upper section) performs robust and low-drift LiDAR odometry. The semantic interpretation and conversion module (lower section) then uses this map to build a voxel map, analyze geometric properties for structural semantic understanding, and produce lightweight geospatial outputs for chart integration.}
  \label{fig:InlandLOAMFramework}
\end{figure*}

The Inland-LOAM framework, as shown in Figure~\ref{fig:InlandLOAMFramework}, is designed to build structural semantic maps of inland waterways for enhanced situational awareness, relying only on LiDAR. The system comprises two primary modules: first, a robust LiDAR odometry and mapping (LOAM) pipeline for accurate pose estimation and initial environmental representation; and second, a semantic interpretation and conversion module that produces enhanced structural semantic maps and lightweight geospatial outputs for integration with standard navigational charts.

In the LOAM module, raw LiDAR point clouds are continuously acquired and preprocessed to remove returns from the ego vessel. The remaining points are projected onto a range image for segmentation. A plane fitting component first estimates the local water plane and labels the corresponding water surface points, while the remaining points are segmented in the range image. The proposed feature extraction method then extracts planar (flat) and edge (rough) features from the processed scan. These feature sets are subsequently used by the two-stage LiDAR odometry component to estimate the vessel pose. The mapping component then refines each pose against a local feature map and incrementally updates the map. Finally, the water plane, odometry, and loop closure constraints are jointly optimized in a factor graph using iSAM2 (incremental smoothing and mapping)~\cite{kaess2012isam2}, as implemented in GTSAM (Georgia Tech Smoothing and Mapping)~\cite{gtsam}.

The output from the LOAM module, comprising the point cloud map and estimated poses, serves as the input for the Conversion module in the lower part of Figure~\ref{fig:InlandLOAMFramework}. In this module, a voxel map is created by dynamically combining features from the LOAM output based on probabilistic occupancy grids \cite{elfes1989using}.
 Next, 3D free and occupied spaces are identified and extracted from this voxel map. Slope estimation is then performed based on the extracted occupied regions. These estimated slopes, combined with local neighborhood features from the voxel map, are subsequently used to determine structural semantic attributes, which collectively form a 2D structural semantic map. Finally, a conversion stage converts this 2D map into lightweight data that can be integrated with standard navigational charts.

\subsection{System Equipment}
\label{subsec3_2}
\begin{figure}[!t]
    \centering
    \includegraphics[width=0.8\columnwidth]{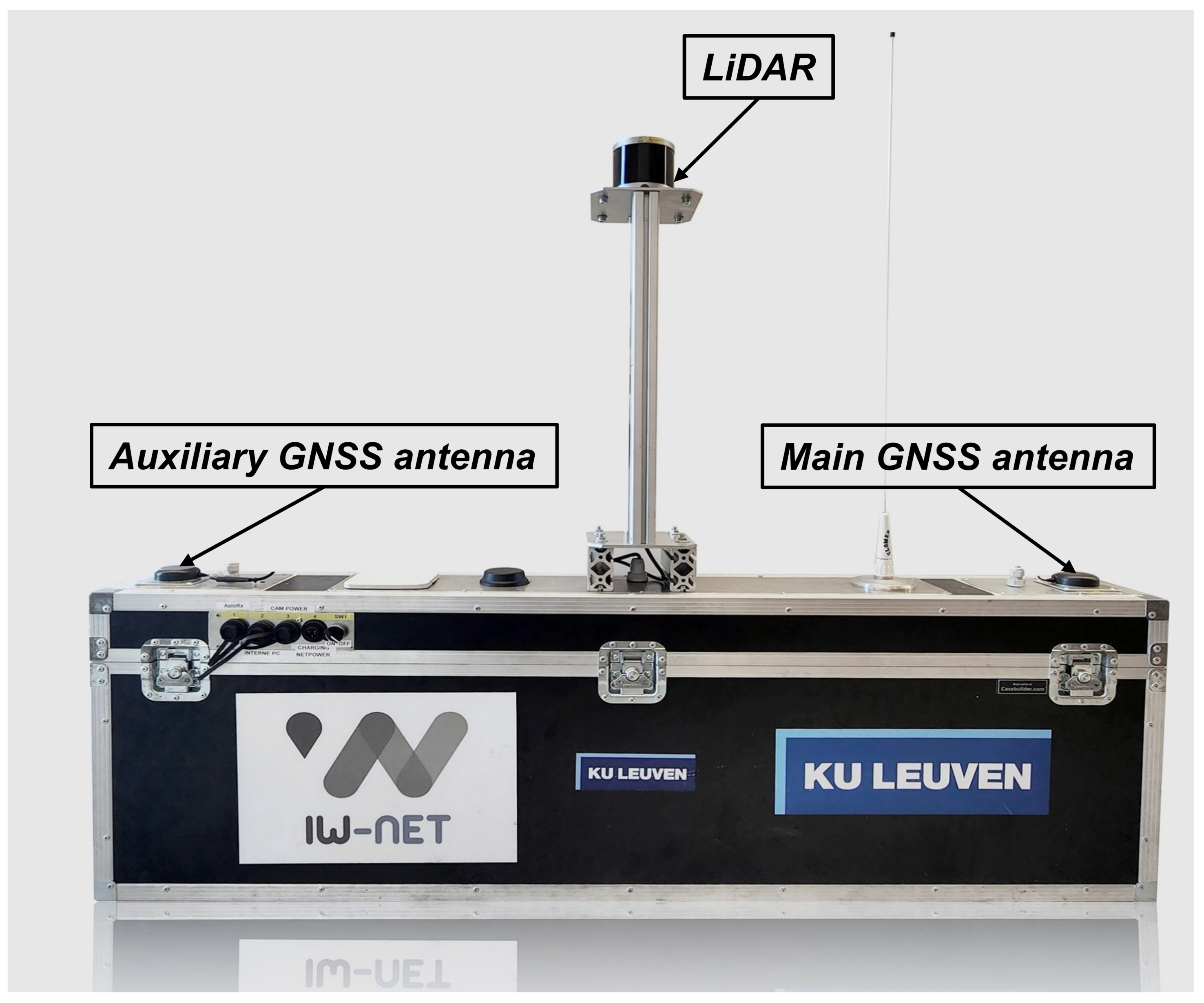}
  \caption{The layout of sensors on our self-developed Sensor Box.}
  \label{fig:SB_fig}
\end{figure}

The data acquisition setup, as illustrated in Figure~\ref{fig:SB_fig}, integrates a RoboSense Helios-32 LiDAR with a Septentrio AsteRx-i3 D Pro+ GNSS/INS receiver.

Regarding the inherent water surface interference challenges of inland waterways discussed in Section~\ref{subsec1}, Allis et al. \cite{allis2011application} demonstrated that optimal detection of water surfaces relies on diffuse reflectivity, influenced predominantly by laser incidence angles and water turbidity. Furthermore, optical constraints, such as grazing angles, inherently limit the detectable surface area. Therefore, a LiDAR with an expanded vertical field of view was selected to maximize the reliability of water surface detection.

The Helios-32 is a 32-line LiDAR with a maximum measurement range of 150 m and an accuracy of $\pm$ 1 cm. It provides a 360-degree horizontal field of view and a 70-degree vertical FoV, characterized by a non-uniform distribution of its scan lines and an approximate vertical angular resolution of 1.33 degrees. In this paper, the LiDAR was operated at a scanning frequency of 10 Hz and a horizontal angular resolution of 0.2 degrees. The LiDAR was mounted with its horizontal scan plane approximately parallel to the water surface; the effective vertical distance from the LiDAR to the water surface was approximately $1.83\,\mathrm{m}$ during data collection.
The AsteRx-i3 D Pro+ GNSS/INS receiver, equipped with dual antennas and an integrated IMU, provides RTK-GNSS/INS measurements at 20 Hz. These measurements are used to identify reliable RTK-fixed intervals and to compute trajectory errors, but they are not used as odometry factors or priors in the proposed estimator.

Data were processed and stored on a laptop equipped with an AMD Ryzen 9 PRO 7940HS CPU and 64 GB of RAM, running Ubuntu 22.04 and ROS 2\footnote{\url{https://www.ros.org/}}.

\subsection{LiDAR Odometry and Mapping Module}
\label{subsec3_3}

\subsubsection{Preprocessing and Plane Fitting}
\label{subsec3_3_1}

The preprocessing begins by converting the raw LiDAR measurements into a structured range image. Let $\mathbf{P}_k = \{\mathbf{p}_i\}_{i=1}^{N}$ represent the LiDAR point cloud collected in the $k$-th scan cycle, where each point $\mathbf{p}_i = [x_i, y_i, z_i]^\transpose \in \mathbb{R}^3$ denotes spatial coordinates. Due to the substantial vertical FoV of the LiDAR and the considerable physical dimensions of the vessel, some of the LiDAR returns inevitably originate from the vessel itself. Therefore, a bounding-box ($\text{bbox}_{\text{vessel}}$) filtering step is applied to exclude these points:
\begin{equation}
\mathbf{P}_k^{\text{valid}} = \{\mathbf{p}_i \mid \mathbf{p}_i \notin \text{bbox}_{\text{vessel}}\}.
\end{equation}

Given the LiDAR's non-uniform vertical distribution of its 32 laser beams and a fixed horizontal resolution of $0.2^\circ$, each scanning cycle produces a 2D range image, $I_k$, of dimension $32 \times 1800$ (rows $\times$ columns). Each row corresponds to an individual laser beam: $v_j$ represents the scan line index (row), while each column represents sequential horizontal angle increments, with $u_j$ denoting the horizontal angle index (column).

For subsequent segmentation, the valid points are stored in a $32\times1800\times3$ information matrix. The three channels contain the 3D coordinates, the binary water surface indicator $M_k^{\mathrm{surface}}$, and the cluster label map $M_k^{\mathrm{label}}$, respectively. Isolated or invalid points are marked as outliers in the label channel.

For short-range localization and mapping, the curvature of the Earth can be neglected, and the water surface within the local LiDAR observation range is approximated as a plane \cite{griesser2023visual}. Candidate water points are selected from a predefined set of lower LiDAR beams according to the LiDAR mounting height and vessel draft. In the implemented configuration, these candidate beams are selected from the downward scan lines, with vertical scan angles approximately ranging from $-55^\circ$ to $-15^\circ$. Instead of using ground extraction based on the angle relationship between adjacent beams~\cite{himmelsbach2010fast}, which can be affected by water surface undulations, we fit the candidates with RANSAC~\cite{fischler1981random}. The water plane in the sensor frame is defined as:

\begin{equation}
    \mathbf{n}_s^\transpose \mathbf{p} + d_s = 0
    \label{eq:water_plane}
\end{equation}

where $\mathbf{n}_s = [a_s, b_s, c_s]^\transpose \in S^2$ is the normal vector of the plane and $d_s\in\mathbb{R}$ is its scalar offset from the sensor origin. Subsequently, any point $\mathbf{p}_j \in \mathbf{P}_k^{\text{valid}}$ is classified as a water surface point, setting the corresponding $M_k^{\mathrm{surface}}(v_j, u_j) = 1$, if its perpendicular distance to this estimated plane, calculated as $|\mathbf{n}_s^\transpose \mathbf{p}_j + d_s|$, is within a predefined tolerance $\tau_s$.

The remaining non-surface points are segmented using range image-based clustering~\cite{bogoslavskyi2016fast}. Starting from each unlabeled point, Breadth-First Search (BFS) groups neighboring pixels when their range and angular relationship do not indicate a strong depth discontinuity, controlled by the threshold $\theta_{\mathrm{seg}}$. Each connected component is assigned a cluster label in $M_k^{\mathrm{label}}$, and segments with fewer than 30 points are relabeled as outliers.

\subsubsection{Feature Extraction}
\label{subsec3_3_2}

\begin{algorithm} [!ht]
\caption{Feature Extraction}
\label{alg:feature_extraction}
\begin{algorithmic}[1]
  \REQUIRE scan: $P$
  \ENSURE feature point sets: edge (rough) $\mathcal{E}$, planar (flat) $\mathcal{P}$
  \item[\textbf{Parameters:}]  thresholds: $\tau_1$, $\tau_2$, $\tau_3$
  \STATE $\mathcal{E} \leftarrow \emptyset,\quad
         \mathcal{P} \leftarrow \emptyset$
  \FORALL{scan line $L_k$ in $P$}
    \FORALL{segment $G_m$ in scan line $L_k$}
        \FORALL{point $p_i$ in segment $G_m$}
    
            \STATE Select neighboring points:
            \STATE $\mathcal{M}_i=\{p_{i-n},\ldots,p_i,\ldots,p_{i+n}\}$
            \STATE Calculate curvature $c_i$ from ranges $\rho_i$:
            \STATE $\displaystyle
              c_i=\left(\frac{1}{2n}
              \sum_{p_j\in\mathcal{M}_i\setminus\{p_i\}}
              (\rho_j-\rho_i)\right)^2$
            \STATE Calculate roughness $r_i$:
            \STATE $\displaystyle
              \bar{\rho}_i=\frac{1}{2n+1}
              \sum_{p_j\in\mathcal{M}_i}\rho_j$
            \STATE $\displaystyle
              r_i=\frac{1}{2n+1}
              \sum_{p_j\in\mathcal{M}_i}(\rho_j-\bar{\rho}_i)^2$
            \IF{$\,c_i>\tau_1 \;\lor\; r_i>\tau_2$}
              \STATE $\mathcal{E}\leftarrow\mathcal{E}\cup\{p_i\}$
            \ELSIF{$r_i<\tau_3$}
              \STATE $\mathcal{P}\leftarrow\mathcal{P}\cup\{p_i\}$
          \ENDIF
        \ENDFOR
    \ENDFOR
  \ENDFOR
  \STATE Exclude unstable points from $\mathcal{E}$, $\mathcal{P}$ caused by occlusion
  \STATE Sort $\mathcal{E}$ and $\mathcal{P}$ within each segment using $c_i$ and $r_i$
  \STATE Non-maximum Suppression on $\mathcal{E}$, $\mathcal{P}$
  \RETURN $\mathcal{E}, \mathcal{P}$
\end{algorithmic}
\end{algorithm}

Our feature extraction process is similar to the method in \cite{shan2018lego}, but with optimized roughness characterization for structural semantic mapping. Features are extracted from the water surface and the non-water segments identified by the preceding segmentation step. A key challenge involves the robust discrimination of surface texture characteristics within local regions.

Traditional approaches often extract features based on local range differences. For instance, a common metric (Equation~\ref{eq:trad_roughness}, where $|L|$ is the size of the local neighborhood, e.g., 10 points, and $\rho_i$ denotes the measured range of point $i$) is:
\begin{equation}
c_i^{\mathrm{range}} =
\frac{1}{|L|\,\rho_i}
\left|
\sum_{j\in L,\;j\neq i}
(\rho_j-\rho_i)
\right|,
\label{eq:trad_roughness}
\end{equation}
While this method is computationally efficient and applicable to all directions, it has limitations in inland waterway environments. In areas with dense vegetation, grass, or gravel, the local surface exhibits numerous small, ``edge-like'' bumps. When applying Equation~\ref{eq:trad_roughness}, the positive and negative range residuals $(\rho_j - \rho_i)$ from neighboring points $p_j$ on either side of the central point $p_i$ tend to cancel each other out. Consequently, such curvature estimations often overlook subtle surface textures and primarily respond only to more extreme geometric features like significant peaks or occlusions.

\begin{figure}[!t]
    \centering
    \includegraphics[width=0.6\columnwidth]{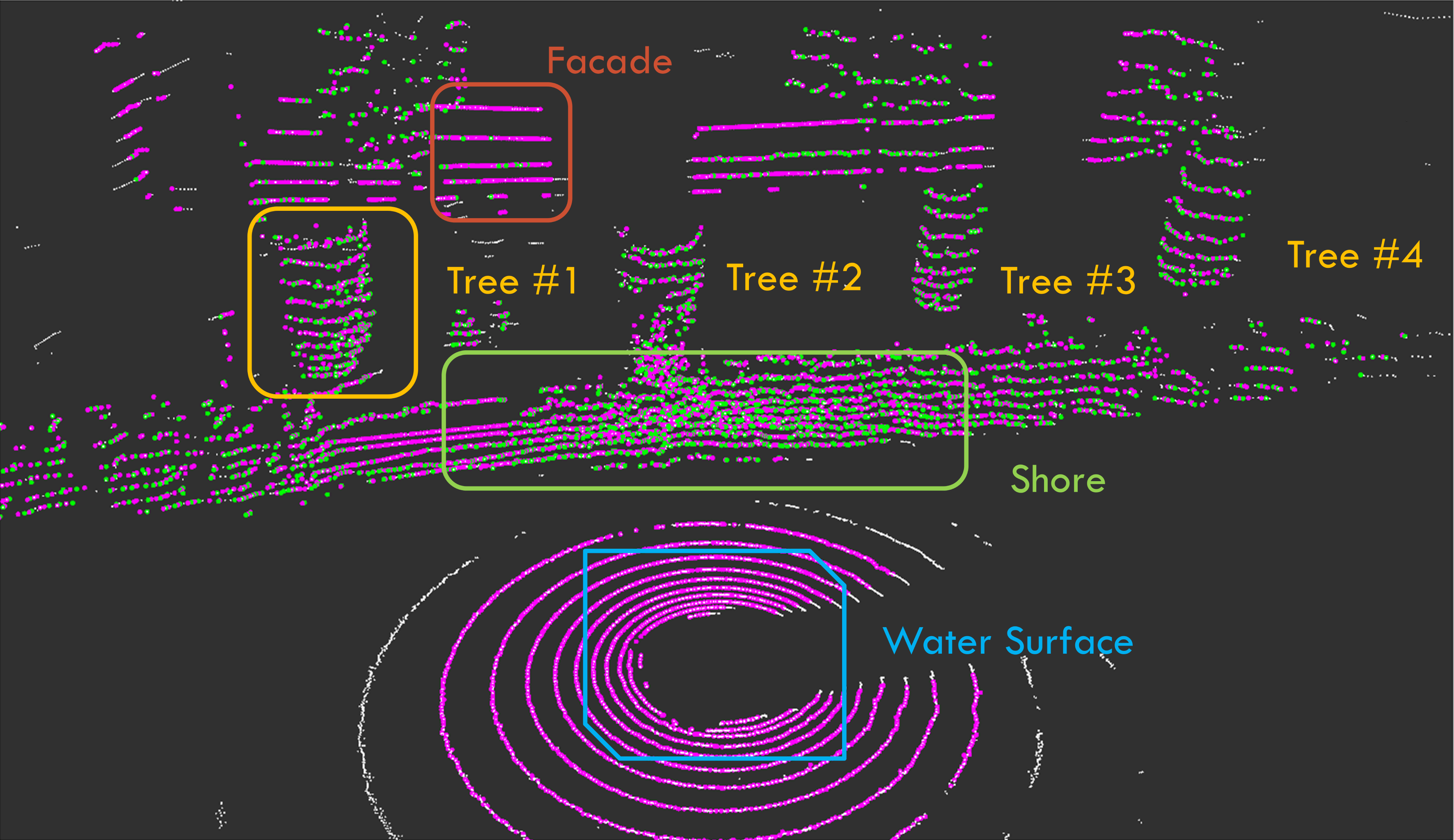}
  \includegraphics[width=0.6\columnwidth]{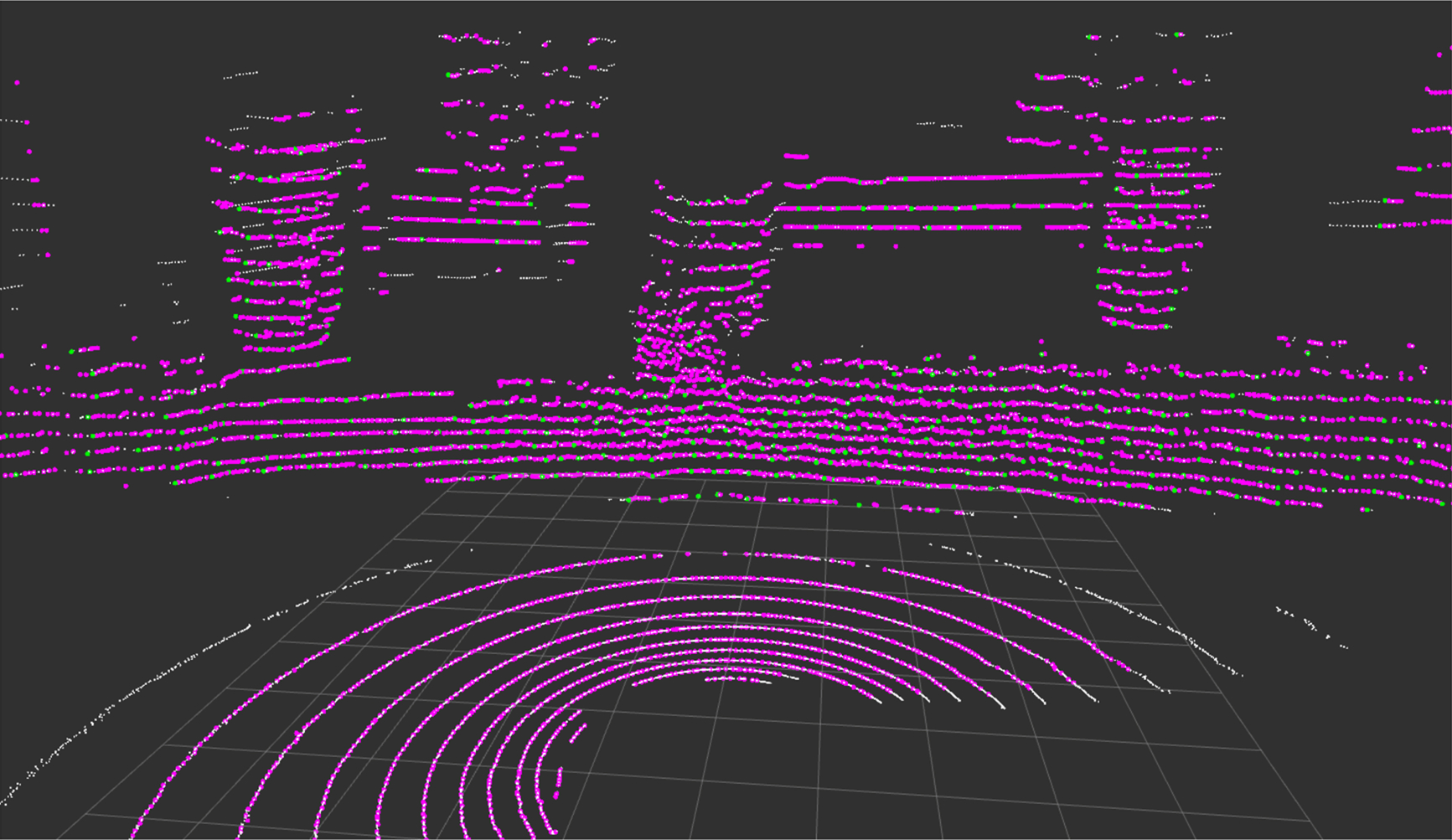}
  \caption{Comparative performance of feature extraction. Purple points indicate planar features, while green points denote rough features. The upper figure illustrates the feature distribution by our method, where the semantic information of the targets is annotated. Notably, the shore is largely covered by vegetation, with only a small portion appearing as a flat surface. The lower figure shows the feature distribution of LOAM-based methods.}

  \label{fig:comparative_feature}
\end{figure}

To overcome this limitation and effectively capture local surface roughness, thereby distinguishing between smooth and textured surfaces, we employ a variance-based approach to quantify local range variations as detailed in Algorithm~\ref{alg:feature_extraction}. Figure~\ref{fig:comparative_feature} illustrates the comparative performance of feature extraction between LOAM-based methods and our optimized approach.

\subsubsection{LiDAR Odometry, Mapping and Joint Optimization}
\label{subsec3_3_3}

\begin{figure}[!t]
  \includegraphics[width=\columnwidth]{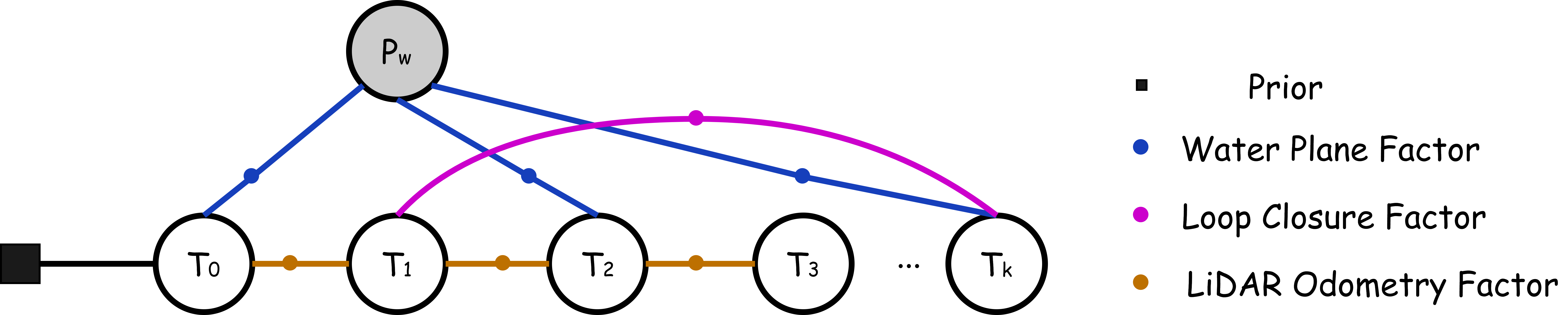}
  \caption{Inland-LOAM factor graph structure. The pose variables $T_k$ are connected by LiDAR odometry and loop closure factors, while water plane factors connect selected poses to the global water surface plane landmark $P_w$. The prior factor anchors the initial state.}
  \label{fig:factor_graph}
\end{figure}

Our LiDAR odometry and mapping modules build upon LeGO-LOAM \cite{shan2018lego}, with specific adaptations in joint optimization. The system employs a two-stage registration process for pose estimation, integrating water surface constraints. The overall factor graph structure is illustrated in Figure~\ref{fig:factor_graph}.

\textbf{Two-Stage LiDAR Odometry:}
The odometry module uses a two-stage Levenberg--Marquardt optimization. Planar features $\mathcal{F}_p^t$, extracted from the water surface and other planar environmental elements, are first matched to $\mathcal{F}_p^{t-1}$ using point-to-plane associations to estimate $[t_z,\theta_{\mathrm{roll}},\theta_{\mathrm{pitch}}]$. Edge features $\mathcal{F}_e^t$ are then matched using point-to-line associations~\cite{zhang2014loam} to estimate $[t_x,t_y,\theta_{\mathrm{yaw}}]$, yielding the full 6-DoF transformation at lower computational cost.

\textbf{LiDAR Mapping:}
The mapping module refines the pose by matching the current feature sets $\{\mathcal{F}_e^t,\mathcal{F}_p^t\}$ to a surrounding feature map. Instead of maintaining a single fused point cloud, we store feature sets with their poses and build the local map from keyframes within the sensor field of view, typically within 100 m. This representation supports later factor graph optimization. However, the resulting LiDAR odometry can still accumulate drift, especially in the vertical direction, where stable horizontal structures are often sparse in many inland waterway environments.

\textbf{Joint Optimization with Water Surface Constraints:}
The water surface provides a global reference plane for reducing drift over long trajectories. Unlike fragmented planar structures in terrestrial scenes, the water surface is broad and continuous in inland waterways. We therefore introduce it as a plane landmark $P_w=(n_w,d_w)$ in the factor graph, as illustrated in Figure~\ref{fig:factor_graph}.

The objective of our estimation is to determine the optimal set of vessel poses $\mathcal{T} = \{T_k\}_{k=0}^{N}$, where $T_k\in SE(3)$.
This is achieved by minimizing a cost function composed of several weighted squared error terms, representing the negative log-likelihood of the measurements given the states and plane parameters. The optimization problem is formulated as finding $\mathcal{X}^* = \{\mathcal{T}^*,P_w^*\}$ by minimizing:

\begin{equation}
\begin{split}
\mathcal{X}^* = \arg\min_{\mathcal{T},P_w}\;&
    \|r_0\|^2_{\Sigma_0}
\\
&+ \sum_k \bigl\|r_{\mathrm{odom},k}(T_k, T_{k+1})\bigr\|^2_{\Sigma_{\mathrm{odom},k}}
\\
&+ \sum_{(i,j)\in\mathcal{L}} \bigl\|r_{\mathrm{loop},ij}(T_i, T_j)\bigr\|^2_{\Sigma_{\mathrm{loop},ij}}
\\
&+ \sum_{m\in\mathcal{M}}
\rho_{\delta}\!\left(
\bigl\|r_{\mathrm{plane},m}(T_m,P_w)\bigr\|^2_{\Sigma_{\mathrm{plane},m}}
\right)
\label{eq:opt}
\end{split}
\end{equation}

Because water plane observations can be affected by reflections, waves, and occasional incorrect plane fits, we apply a Huber robust noise model to the water plane residuals. Let
$s_m=\|r_{\mathrm{plane},m}(T_m,P_w)\|^2_{\Sigma_{\mathrm{plane},m}}$.
The robust cost is defined as
\[
\rho_{\delta}(s_m)=
\begin{cases}
s_m, & s_m \leq \delta^2,\\
2\delta\sqrt{s_m}-\delta^2, & s_m > \delta^2.
\end{cases}
\]
Thus, small water plane residuals are treated as least-squares, while large residuals are downweighted. This formulation preserves the factor graph structure and only reduces the influence of unreliable water plane observations.

In Equation~\eqref{eq:opt}, $\|r\|^2_{\Sigma}$ denotes the squared Mahalanobis distance $r^\transpose\Sigma^{-1}r$, where $\Sigma$ is the covariance matrix of the corresponding residual $r$. The term $r_0$ represents a combined prior residual on the initial pose $T_0$ or the initial plane estimate, with covariance $\Sigma_0$. The odometry and loop closure residuals are defined as

\begin{equation}
\begin{aligned}
r_{\mathrm{odom},k} &=
\mathrm{Log}\bigl((\Delta\widetilde T_{k,k+1})^{-1}\,(T_k^{-1}T_{k+1})\bigr)\in\mathbb{R}^6,\\
r_{\mathrm{loop},ij} &=
\mathrm{Log}\bigl((\Delta\widetilde T_{i,j})^{-1}\,(T_i^{-1}T_j)\bigr)\in\mathbb{R}^6,\quad (i,j)\in\mathcal{L}.
\end{aligned}
\end{equation}

Here, $\mathrm{Log}(\cdot)$ maps an $SE(3)$ element to its tangent space representation, and $\mathcal{L}$ denotes the set of loop closures. The loop closure transform $\Delta\widetilde T_{i,j}$ is calculated using ICP~\cite{besl1992method} and added to a pose graph optimized by iSAM2~\cite{kaess2012isam2}; in the experiments, this module is activated only for the loop-back trajectory in Area A. The sum $\sum_{m\in\mathcal{M}}$ accounts for all water surface plane observations, where $\mathcal{M}$ is the set of keyframes with such observations.

The water plane residual, $r_{\mathrm{plane},m}(T_m,P_w)$, is built from a local water plane observation $P_{s,m}=(n_{s,m},d_{s,m})$ at keyframe $m$ with pose $T_m=(R_m,t_m)$. Using the plane convention $n^\transpose x+d=0$ and the pose convention $x_w=R_mx_s+t_m$, this local observation is transformed into the global frame as $\widehat P_{w,m}=(\hat n_{w,m},\hat d_{w,m})$, where
\begin{equation}
\begin{aligned}
\hat n_{w,m} &= R_m n_{s,m},\\
\hat d_{w,m} &= d_{s,m}-\hat n_{w,m}^{\transpose}t_m .
\end{aligned}
\end{equation}
The residual then compares the estimated global plane $P_w=(n_w,d_w)$ with this transformed observation. Using the plane error operator $\ominus$ between $P_w$ and $\widehat P_{w,m}$, the residual is written as

\begin{equation}
\begin{split}
r_{\mathrm{plane},m}(T_m,P_w)
  &= P_w \ominus \widehat P_{w,m} \\
  &= 
     \begin{pmatrix}
       e_N\!\bigl(n_w,\hat n_{w,m}\bigr) \\
       d_w - \hat d_{w,m}
     \end{pmatrix} \\
  &=
     \begin{pmatrix}
       e_N\!\bigl(n_w, R_m n_{s,m}\bigr) \\
       d_w - \bigl(d_{s,m} - (R_m n_{s,m})^{\mskip2mu\transpose} t_m\bigr)
     \end{pmatrix}
     \in\mathbb{R}^3 .
\end{split}
\end{equation}

Here, $e_N(n_w,\hat n_{w,m})\in\mathbb{R}^2$ is the 2D error vector representing the normal difference between $n_w$ and $\hat n_{w,m}$. It is computed by first defining a 3D tangent space error $\hat{\xi}$ at $n_w$ and then projecting it onto the orthonormal basis $B_{n_w}\in\mathbb{R}^{3\times2}$:

\begin{equation}
\begin{aligned}
&\hat\xi = -\frac{\arccos(n_w^\transpose\hat n_{w,m})}{\sqrt{1-(n_w^\transpose\hat n_{w,m})^2}} \, (\hat n_{w,m} - (n_w^\transpose\hat n_{w,m})\,n_w),\\
&e_N(n_w,\hat n_{w,m})
= B_{n_w}^{\transpose}\hat\xi .
\end{aligned}
\end{equation}

The columns of $B_{n_w}$ span the 2D tangent space at $n_w$ on the unit sphere $S^2$. 

By leveraging the water surface as a global geometric constraint, the proposed formulation provides an additional reference for reducing vertical drift in weakly structured waterway scenes.

\subsection{Structural Semantic Map Conversion}
\subsubsection{Semantic Voxel Transformation}
\label{subsec3_4_1}

The structured point cloud generated by the preceding module contains semantic labels such as water surface points, flat points, and rough points. To store and query this information efficiently, we transform the annotated point cloud into a sparse semantic voxel grid. For each voxel, semantic evidence is fused over time using a log-odds update mechanism adapted from probabilistic occupancy grids~\cite{elfes1989using}.

The workspace is discretized into a 3D voxel grid with a uniform resolution $r=0.3\,\mathrm{m}$, which balances memory use, computational cost, and the scale of inland waterway structures. For each voxel $v_i$, a hash map stores its semantic state $S(v_i)$ over four voxel states: water, flat, rough, and unknown.

\begin{figure}[!t]
  \centering
  \includegraphics[width=\columnwidth]{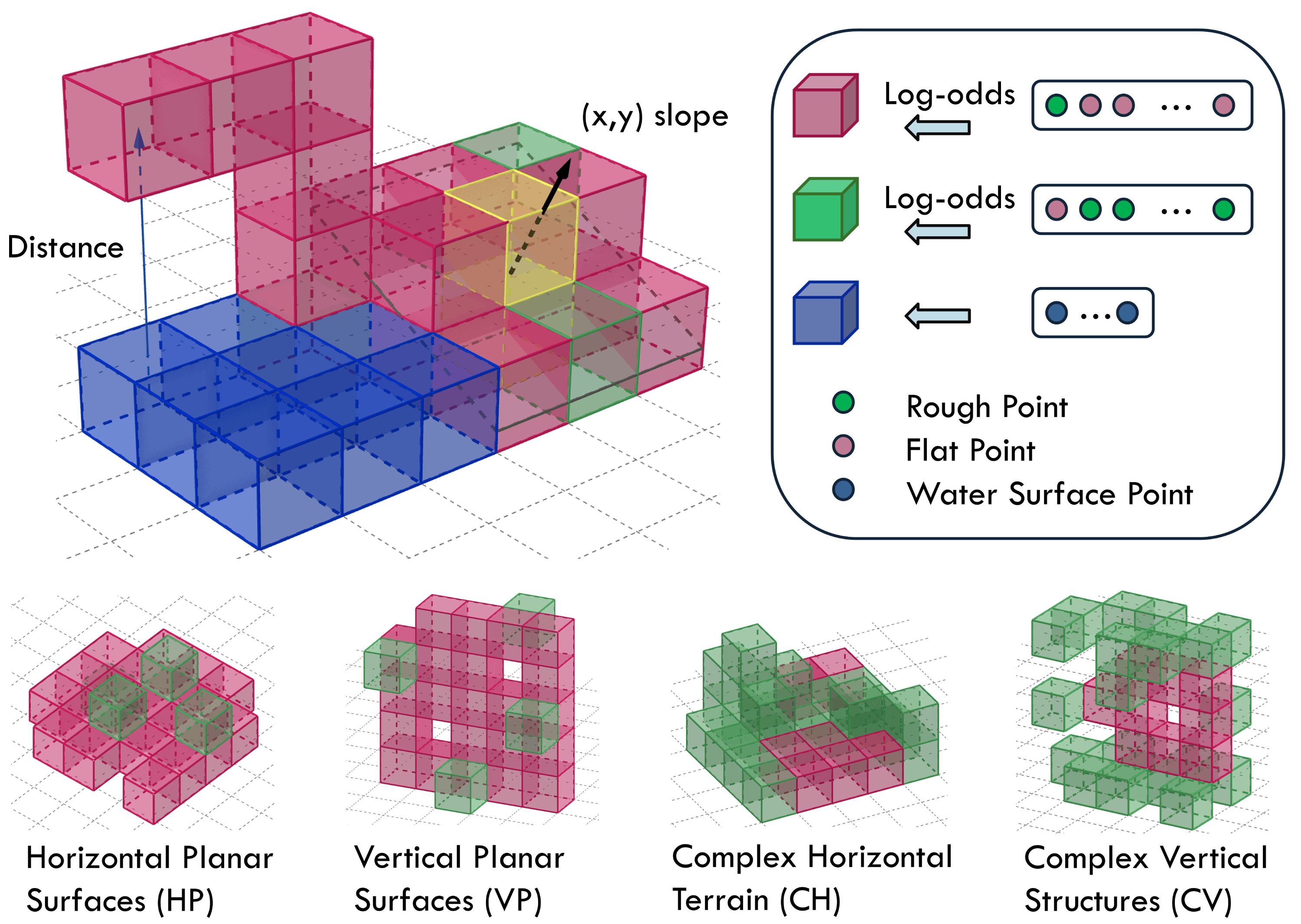}
    \caption{Overview of the structural semantic map conversion. Red and blue voxels denote non-water and water regions, respectively. The voxel map is then used for slope estimation and bridge clearance calculation: the yellow voxel denotes the target cell for local slope fitting, and bridge clearance is measured as the vertical distance from water voxels to the non-water voxels above. The four structural semantic classes at the bottom are generated using a local window with $S_a=2$.}
  \label{fig:voxel_transformation}
\end{figure}

Points pre-classified as water surface by the LiDAR odometry module are handled directly: if such a point falls into voxel $v_i$, the voxel is immediately assigned the ``water'' semantic label with high confidence. 

For terrain characteristics, we maintain log-odds values within the hash map entry for each non-water voxel. The initial belief for each of these log-odds components in every newly considered voxel is denoted as $l_{0, i}$.

When a new semantic point $p_t$ belonging to a terrain class $c \in \{\text{flat, rough}\}$ is observed at time $t$ and falls within voxel $v_i$, the log-odds value for that class in the voxel is updated as follows:

\begin{equation}
l_{t,i}^c = \text{logit}(p(m^c | y_t)) + l_{t-1,i}^c - l_{0,i}^c
\end{equation}
where the log-odds in voxel $v_i$ at time $t$ is $l_{t,i}^c$. The term $p(m^c | y_t)$ represents the probability of observing class $c$ given the measurement $y_t$. The logit function is defined as $\text{logit}(p) = \ln\left(\frac{p}{1-p}\right)$. We define the observation model probabilities for semantic classes, $p(\text{flat}) = p_f$ (e.g., $p_f = 0.7$) and $p(\text{rough}) = 1 - p_f$.

Following the update, the log-odds $l_i^c$ for each voxel $v_i$ and each class $c$ are converted back to probabilities:

\begin{equation}
P_i^c = \sigma(l_i^c) = \frac{1}{1 + e^{-l_i^c}}
\end{equation}

Ultimately, each voxel $v_i$ is assigned a definitive semantic label. The final label \(S(v_i)\) is determined from \(P_i^{\mathrm{flat}}\) as follows:

$S(v_i) =
\begin{cases}
\text{``flat''} & \text{if } P_i^{\text{flat}} - 0.5 > \tau_{\text{flat}} \\
\text{``rough''} & \text{if } P_i^{\text{flat}} - 0.5 < -\tau_{\text{rough}} \\
\text{``unknown''} & \text{otherwise}
\end{cases}$

\subsubsection{2D Structural Semantic Map Construction}
\label{subsec3_4_2}

The 3D semantic voxel map is subsequently projected onto a unified 2D structural semantic map. This projection builds on the conversion method in~\cite{fredriksson2024voxel}, but is adapted to USV scenarios with refined slope estimation and structural semantic inference.

To represent the complex environments encountered by USVs in inland waterways, the non-water regions are divided into four structural semantic classes: \textit{Horizontal Planar Surfaces} (HP), including level terrain and flat artificial or natural shore platforms; \textit{Vertical Planar Surfaces} (VP), including building facades, bridge structures, vessels, and vertical embankments; \textit{Complex Horizontal Terrain} (CH), referring to irregular embankments, grasslands, and low-lying scrub; and \textit{Complex Vertical Structures} (CV), representing non-planar upright natural elements such as trees and dense bankside vegetation. Figure~\ref{fig:voxel_transformation} visualizes representative examples of these four classes. Two additional classes are defined for water-related regions: \textit{Open Water Surface} (OW) and \textit{Under-Bridge Clearance Zone} (UB), which represent open water and navigable water passages beneath bridge structures, respectively.

The foundation of this conversion is an intermediate 2D grid representation, where each cell $(x, y)$ retains detailed information regarding the vertical distribution of voxels and their semantic labels derived from the 3D voxel map. Specifically, for each cell $(x, y)$ at the defined map resolution, we consider the voxel set:
$
V_{xy} = \bigl\{(z_j,\, s_j,\, p^{\mathrm{orig}}_j)\bigr\}_{j = 1}^{N_{xy}}
$
where $z_j$ is the voxel height, $s_j$ is the semantic label (e.g., water, flat, rough, and unknown), $p^{\mathrm{orig}}_j$ denotes the original 3D coordinates, and $N_{xy}$ is the total number of voxels in the vertical column at (x, y). For outdoor environments lacking explicit ceilings, the local maximum voxel height $Z_{max}$ serves as an initial estimate for the upper boundary of free space above newly detected occupied regions. 

The next step is to perform slope estimation on these voxel sets, as illustrated in Figure~\ref{fig:voxel_transformation}. In contrast to methods that derive slope using only boundary heights, our approach calculates the slope through all voxel points within a defined local neighborhood. This method is more suitable for estimating structures such as VP, CV, or overhangs.

For a target cell $(x_c,y_c)$, we define a square neighborhood window
$W_{xy}$ of size $(2S_a + 1) \times (2S_a + 1)$, where $S_a$ specifies the extent of the neighborhood.

Each voxel in $W_{xy}$ is represented by its relative grid coordinates and
height $(x_k^{\mathrm{rel}},\,y_k^{\mathrm{rel}},\,z_k)$, measured with
respect to $(x_c,y_c)$, where $(x_k^{\mathrm{rel}})$ and $(y_k^{\mathrm{rel}})$ are expressed in grid-cell units. Let $N$ be the total number of such voxel data points aggregated from all cells within the entire window $W_{xy}$. The slope estimation is then performed using a least-squares fit of a plane. The plane fitting is formulated as:

\begin{equation} \label{eq:matrix_slope_fit_detailed}
\underbrace{
\begin{bmatrix}
x^{\mathrm{rel}}_{1} & y^{\mathrm{rel}}_{1} & 1 \\
x^{\mathrm{rel}}_{2} & y^{\mathrm{rel}}_{2} & 1 \\
\vdots & \vdots & \vdots \\
x^{\mathrm{rel}}_{N} & y^{\mathrm{rel}}_{N} & 1
\end{bmatrix}
}_{\textstyle A}
\underbrace{
\begin{bmatrix} a \\ b \\ c \end{bmatrix}
}_{\textstyle \theta}
=
\underbrace{
\begin{bmatrix} z_{1} \\ z_{2} \\ \vdots \\ z_{N} \end{bmatrix}
}_{\textstyle B}.
\end{equation}

The least-squares parameters are obtained as
$\theta_{\mathrm{ls}}=[a_{\mathrm{ls}},b_{\mathrm{ls}},c_{\mathrm{ls}}]^\transpose=A^+B$,
where $A^+$ is the Moore--Penrose pseudoinverse of $A$ computed by singular value decomposition (SVD) for numerical stability. The slope at the central cell $(x_c,y_c)$ is then computed as:
\begin{equation}
m_{xy}= \sqrt{\,a_{\mathrm{ls}}^{2}+b_{\mathrm{ls}}^{2}} .
\end{equation}

The structural label for each 2D grid cell is determined by comparing three cues---$N$ (the number of reliable voxels in the window); $R_{\mathrm{plane}}$ (the fraction of those voxels that belong to locally planar surfaces $R_{\mathrm{plane}} \;=\; \frac{N_{\mathrm{plane}}}{N}$); and the local slope $m_{xy}$.
Let $\{N_{\mathrm{low}}, N_{\mathrm{high}}\}$, $\{R_{\mathrm{low}}, R_{\mathrm{high}}\}$, and $\{m_{\mathrm{low}}, m_{\mathrm{high}}\}$ denote the corresponding thresholds. A cell is classified as:
\begin{itemize}
    \item \textbf{HP} if $N > N_{\mathrm{high}}$, $R_{\mathrm{plane}} > R_{\mathrm{high}}$ and $m_{xy} < m_{\mathrm{low}}$;
    \item \textbf{VP} if $N > N_{\mathrm{low}}$, $R_{\mathrm{plane}} > R_{\mathrm{low}}$ and $m_{xy} > m_{\mathrm{high}}$;
    \item \textbf{CH/CV} if neither the HP nor VP conditions are met, with the class determined by the local slope;
\end{itemize}

A grid cell that contains only voxels labeled as water is assigned to \textit{Open Water Surface} (OW). If a non-water voxel layer appears above the water layer by more than the safety margin $\tau_{\mathrm{bridge}}$ (set to $3.0\,\mathrm{m}$ in this work), the cell is assigned to \textit{Under-Bridge Clearance Zone} (UB). The bridge clearance is computed as $G_{\mathrm{bridge}} = Z_{\text{overhead\_lowest}} - Z_{\text{water\_level}}$, where $Z_{\text{overhead\_lowest}}$ is the height of the lowest overhead structure and $Z_{\text{water\_level}}$ is the height of the water surface in the same vertical column. These additional rules override the geometric label whenever they trigger, reducing the likelihood of misclassification.

 To provide a more practical representation of the clearance under the bridge, we first divide the 2D grid map into local blocks of size $l \times l$ and compute the mean and range of $G_{\mathrm{bridge}}$ within each block. These blocks are then further grouped into larger contiguous regions using BFS. For each region, the mean clearance of its constituent blocks is taken as the representative value, so that each under-bridge area is assigned a single vertical clearance. This two-stage aggregation ensures both the accuracy and real-time performance of the measurement.

\subsubsection{Generation of Map Data for Standard Navigational Charts}
\label{subsec3_4_3}

This process transforms the 2D structural semantic map into a durable and shareable representation for compatibility with IENC standards. It extracts two primary feature types from the structural semantic map: shoreline points and bridge polygons. The method first encodes the relevant occupancy classes into an RGB image, where each class is assigned a unique color value.

For \textbf{Shoreline Extraction Represented by Points}, a ray-casting method is used to delineate shorelines. From the boundaries of identified water regions, rays are projected outward in multiple directions. The end points of these rays, where they intersect the nearest non-water pixels, define the shoreline.

For \textbf{Bridge Extraction Represented by Polygons}, bridge pixels are first identified and then clustered using DBSCAN~\cite{ester1996density} to group fragmented detections into coherent structures. For each cluster, a convex hull is computed to generate a polygonal representation. Overlapping or closely spaced polygons are merged to ensure continuous bridge structures are represented as unified entities. The resulting polygons are exported in JSON format so that they can be used by navigation and chart update modules without changing the IENC chart structure.

The exported layer is not a complete native IENC product. Instead, it is a georeferenced map overlay containing shoreline points, bridge polygons, estimated clearance attributes, timestamps, and source metadata. In an IWT navigation system, this overlay can support local path planning and bridge clearance checking during operation. It can also provide recent local evidence for offline chart maintenance. In this way, existing IENC objects remain the authoritative chart layer, while the Inland-LOAM output serves as a supplementary source for validation and update.

\section{Experiments and Discussion}
\label{sec4}

\subsection{Data Collection}
\label{sec4_1}

To validate Inland-LOAM, we collected one dataset comprising five routes, denoted Areas A--E, along the canal from Leuven to the Dijle (50$^\circ$53.8890$^\prime$N, 004$^\circ$42.4010$^\prime$E) using the Maverick catamaran shown in Figure~\ref{fig:Maverick}. Unlike open lakes, estuaries, or wide rivers, the test routes are located in an inland canal where the effective wind fetch is limited by banks, bends, bridges, locks, and surrounding infrastructure. Therefore, the experiments evaluate the proposed water surface constraint under confined canal conditions with low to moderate wind and traffic influence, rather than under open-water wave conditions. 

The catamaran has an overall length of $6.10\,\mathrm{m}$ and a beam of $2.02\,\mathrm{m}$, with a draft varying from approximately $0.30\,\mathrm{m}$ in lightship condition to $0.60\,\mathrm{m}$ when fully ballasted. Its twin-hull configuration provides stable motion for sensor data acquisition, and the vessel was equipped with the sensor suite described in Section~\ref{subsec3_2}. During data collection, the catamaran operated at speeds up to approximately $2\,\mathrm{m/s}$, covering both slow maneuvering near bridges and faster traversal along open canal sections. The water turbidity measured near the acquisition site ranged from 35 to 56 NTU, indicating turbid water conditions similar to those discussed by Allis et al.~\cite{allis2011application} for water surface recognition using LiDAR. Under these conditions, stable LiDAR returns from the water surface were observed and used for water plane extraction.

\begin{figure}[!t]
  \centering
  \includegraphics[width=0.9\columnwidth]{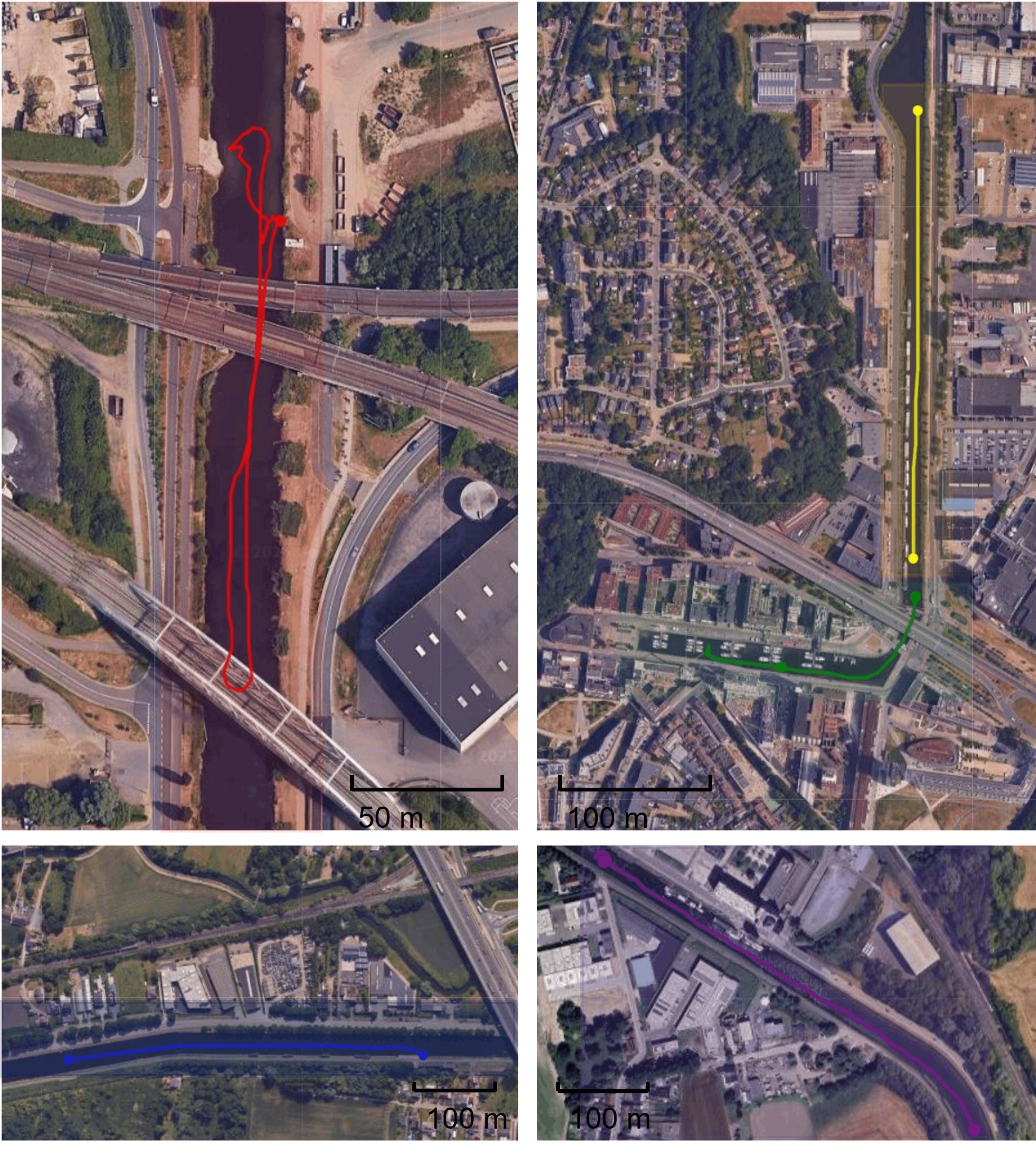}
  \caption{Our data collection was conducted along the canal from Leuven to the Dijle. The collected trajectories are visualized as follows: Area A in red, B in yellow, C in green, D in blue, and E in purple.}
  \label{fig:traj_google}
\end{figure}

As shown in Figure~\ref{fig:traj_google} and Table~\ref{tab:trajectorydetails}, the five routes cover mixed shorelines, vegetation-dominated banks, and an urban port environment, providing representative inland waterway scenes for evaluating localization and semantic mapping.

\begin{itemize}
    \item \textbf{Mixed Environments (Areas A and E)}: These areas contain diverse bank structures, including quay walls, stone revetments, vegetated banks, and exposed soil. Area A includes two fixed bridges. The Area E route does not pass beneath any bridges but includes nearby buildings and several docked vessels.
    \item \textbf{Vegetation Environments (Areas B and D)}: These areas are dominated by densely vegetated banks. Area B includes several mooring facilities and multiple docked vessels, while Area D represents a more unstructured natural environment with a docking platform.
    \item \textbf{Urban Port Environment (Area C)}: This area is characterized by large buildings on both sides of the port and includes a fixed bridge, a lifting bridge, multiple mooring facilities, and small docked boats.
\end{itemize}

\begin{table}[htbp] % Use table* for two-column span
  \centering
  \caption{Evaluated inland waterway areas.} 
  \label{tab:trajectorydetails}
  \setlength{\tabcolsep}{2pt}
  \footnotesize 
  \begin{tabular}{lcccc}
    \toprule
    Area & Distance [m] & Duration [s] & Loop-back & Weather \\
    \midrule
    A (Mixed)      & 424.5 & 561.7 & \ding{51} & Sunny      \\
    B (Vegetation) & 583.5 & 356.8 & \ding{55} & Overcast  \\
    C (Urban Port) & 474.9 & 399.9 & \ding{55} & Overcast  \\
    D (Vegetation) & 527.6  & 436.7  & \ding{55}      & Sunny      \\ 
    E (Mixed)      & 526.3  & 424.0  & \ding{55}      & Overcast      \\ 
    \bottomrule
  \end{tabular}
\end{table}

\subsection{Evaluation of Trajectory and Point Cloud Mapping}
\label{sec4_2}

To evaluate the performance of our method, we compared it with four state-of-the-art methods: KISS-ICP, HDL-SLAM, LeGO-LOAM, and LOAM. To ensure a fair and consistent comparison, all methods were adapted to the laser-beam distribution of the sensor used to collect the dataset. For HDL-SLAM specifically, we used the odometry-only mode with planar constraints enabled.

\begin{figure*}[!t]
  \centering
  \includegraphics[width=0.8\textwidth]{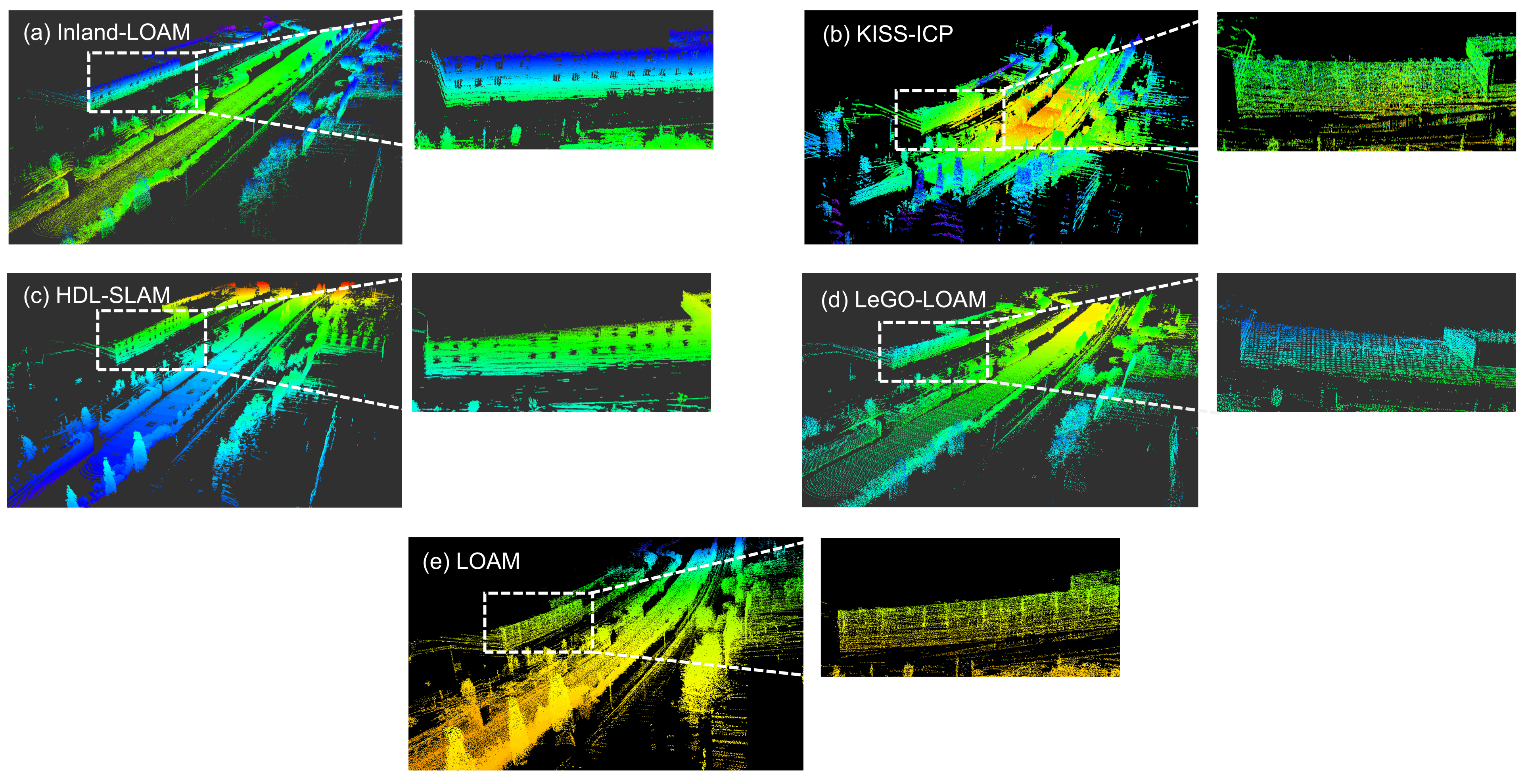}
  \caption{Mapping results of Inland-LOAM, KISS-ICP, HDL-SLAM, LeGO-LOAM, and LOAM in the vegetation area, with a building detail by each method in this scene. The gradient coloring in the Z-axis direction indicates elevation changes.}
  \label{fig:mapping_results}
\end{figure*}

To assess the quality of the generated point cloud maps, we visually compared Inland-LOAM with these baseline methods within Area B, as illustrated in Figure~\ref{fig:mapping_results}. Because Area B has no bridges or large overhead structures, vertical localization is challenging there.

At a global scale, LOAM and KISS-ICP show a water surface that curves downward along the Z-axis, while HDL-SLAM exhibits an upward curvature. In contrast, LeGO-LOAM displays only minor Z-axis drift. Inland-LOAM maintains a relatively flat water surface, demonstrating superior mitigation of vertical drift. At a local scale, Figure~\ref{fig:mapping_results} details the mapping quality of the same building produced by each method, reflecting the precision of their horizontal registration. KISS-ICP, LOAM, and LeGO-LOAM fail to reconstruct the building's windows, capturing only the main facade. While HDL-SLAM correctly identifies the windows, it fails to represent their proper alignment. However, Inland-LOAM accurately reconstructs both the windows and their correct spatial arrangement, reflecting the improved registration of our method.

\begin{figure*}[!t]
  \centering
  \includegraphics[width=0.85\textwidth]{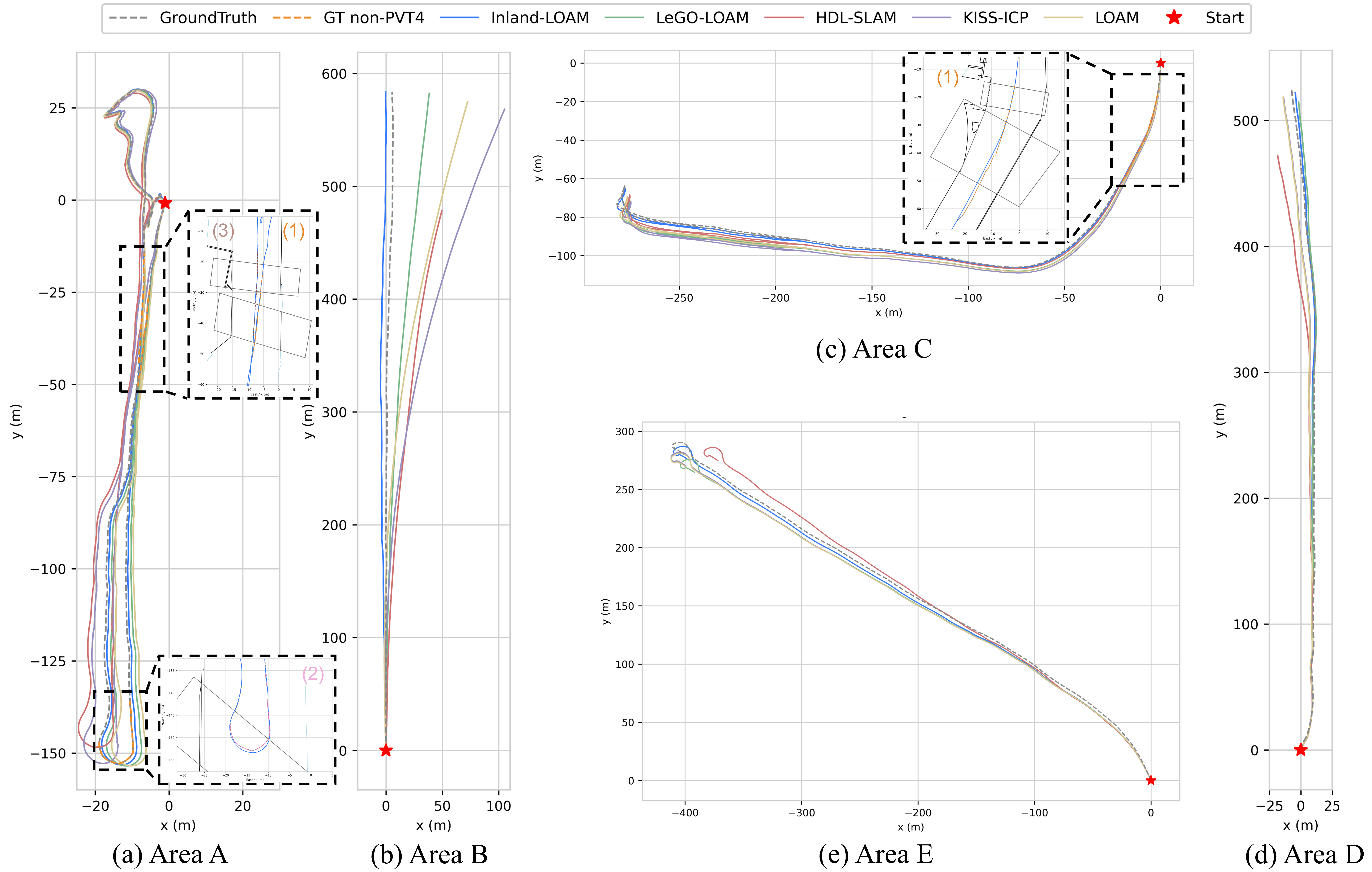}
  \caption{Comparison of the trajectories estimated by all evaluated algorithms with the RTK-GNSS reference trajectory. The plots show the trajectories in the horizontal plane, with each estimate aligned with the reference at its origin. In subfigures (a) and (c), the numbered colored segments identify the GNSS-degraded intervals ($A_{I-1}$, $A_{I-2}$, $A_{I-3}$, and $C_{I-1}$) used for the trajectory re-entry evaluation. The corresponding trajectory segments are projected onto the IENC reference map; in these local views, Inland-LOAM (blue) is rigidly aligned with the RTK-GNSS reference (black) for visual comparison.}
  \label{fig:mapping_traj}
\end{figure*}

The trajectory estimation accuracy of all evaluated methods was quantitatively assessed, with the comparative trajectories presented in Figure~\ref{fig:mapping_traj}. We calculated the Root Mean Square Error (RMSE) and Standard Deviation (STD) between the trajectories generated by these methods and the RTK-GNSS positions, as summarized in Table~\ref{tab:trajectory_accuracy}. Because GNSS signals can drift in inland waterway scenarios, particularly under bridges and in narrow urban sections in Areas A and C, the reliability of the reference trajectory was determined using the Position, Velocity, and Time (PVT) solution status recorded by the receiver. Only the highly reliable RTK-fixed measurements (\texttt{pvt\_mode=4}, indicating fixed ambiguities) were retained for the RMSE and STD calculations; non-fixed samples in Areas A and C were excluded from the accuracy evaluation.

\begin{figure*}[!t]
  \centering
  \includegraphics[width=0.90\textwidth]{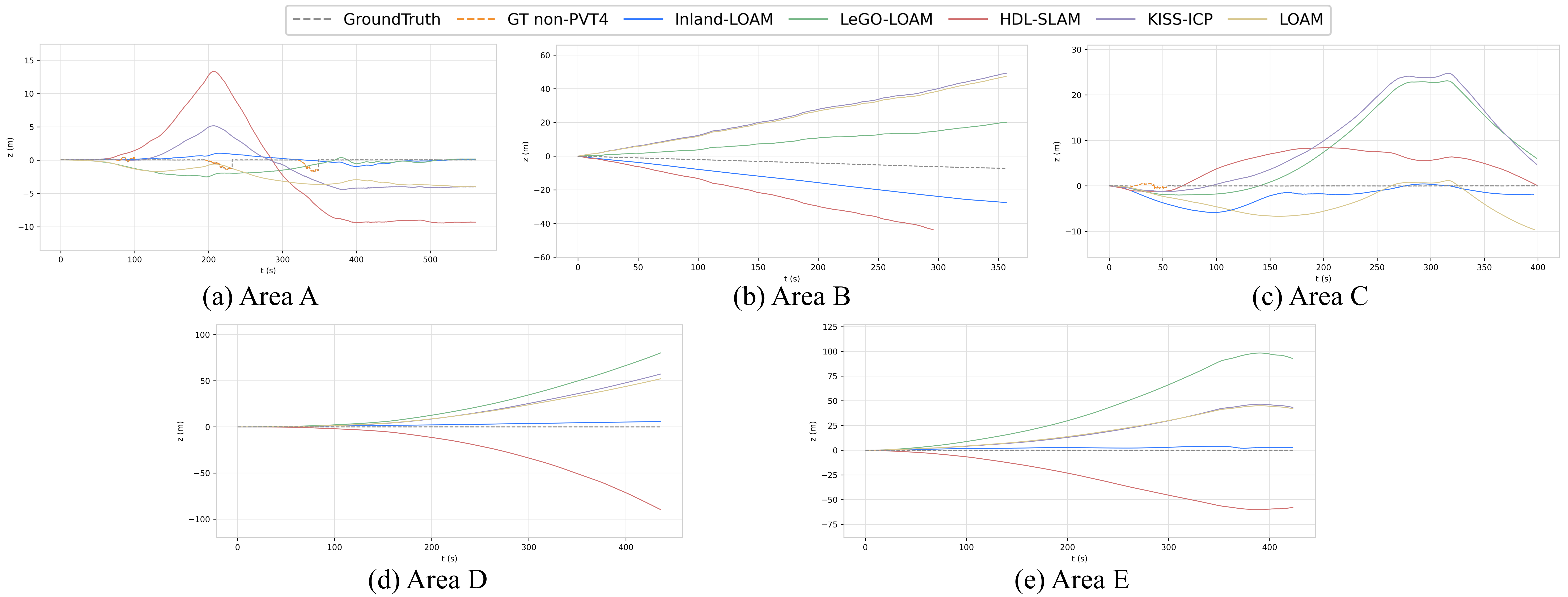}
  \caption{Vertical trajectory errors of all evaluated algorithms relative to the RTK-GNSS reference trajectory.}
  \label{fig:mapping_traj_z}
\end{figure*}

As shown for Areas B, D, and E in Figure~\ref{fig:mapping_traj}, the Inland-LOAM trajectory closely aligns with the RTK-GNSS ground truth, while other methods exhibit a noticeable gap. In Areas A and C, HDL-SLAM and KISS-ICP display larger deviations, while Inland-LOAM consistently maintains a small deviation from the reliable RTK-fixed reference. Regarding the Z-axis comparison in Figure~\ref{fig:mapping_traj_z}, Inland-LOAM demonstrates the lowest error in all scenarios. As shown in Table~\ref{tab:trajectory_accuracy}, Inland-LOAM achieves the lowest RMSE among the configurations without loop closure in each test area. For Area A, however, activating loop closure increases the RMSE from 0.09 m to 0.10 m, while the STD decreases from 0.05 m to 0.03 m. The highly similar scenery in narrow waterways can make loop candidates ambiguous, and erroneous loop constraints may outweigh the benefit of correcting the small accumulated odometry drift. This result highlights a limitation of loop closure in repetitive waterway environments.

To further evaluate the local odometry consistency within these excluded GNSS-degraded intervals shown in Figure~\ref{fig:mapping_traj}, we introduce a trajectory re-entry metric, with the results summarized in Table~\ref{tab:gap_closure}. Specifically, for each continuous non-fixed GNSS interval in Areas A and C, the estimated trajectory is aligned using the last reliable RTK-fixed sample before the degradation, and the 3D position error is computed at the first recovered RTK-fixed sample after the interval.

Table~\ref{tab:gap_closure} shows that Inland-LOAM maintains low re-entry errors in all three intervals of Area A, indicating stable local odometry through bridge and narrow canal passages where GNSS is unreliable. In interval $C_{I-1}$, LOAM obtains the smallest re-entry error of 0.23 m, compared with 0.61 m for Inland-LOAM.

\begin{table*}[htbp]
\footnotesize
\setlength{\tabcolsep}{3.5pt}
\centering
\caption{Algorithm trajectory accuracy evaluation (Unit: m).}
\label{tab:trajectory_accuracy}
\begin{tabular}{l*{16}{c}} 
\toprule
 & \multicolumn{2}{c}{Inland-LOAM}
 & \multicolumn{2}{c}{Inland-LOAM*}
 & \multicolumn{2}{c}{KISS-ICP}
 & \multicolumn{2}{c}{HDL-SLAM}
 & \multicolumn{2}{c}{HDL-SLAM*} 
 & \multicolumn{2}{c}{LeGO-LOAM}
 & \multicolumn{2}{c}{LeGO-LOAM*}
 & \multicolumn{2}{c}{LOAM} \\
 
\cmidrule(lr){2-3}
\cmidrule(lr){4-5}
\cmidrule(lr){6-7}
\cmidrule(lr){8-9}
\cmidrule(lr){10-11}
\cmidrule(lr){12-13}
\cmidrule(lr){14-15}
\cmidrule(lr){16-17}

Area 
 & RMSE & STD
 & RMSE & STD
 & RMSE & STD
 & RMSE & STD
 & RMSE & STD
 & RMSE & STD
 & RMSE & STD
 & RMSE & STD \\
\midrule

A & \textbf{0.09} & 0.05
  & 0.10 & \textbf{0.03} 
  & 1.28 & 0.72 
  & 3.18 & 1.42 
  & 3.19 & 1.43 
  & 0.14 & 0.08 
  & 0.45 & 0.34 
  & 1.32 & 0.59 \\

B & \textbf{0.58} & \textbf{0.30} 
  & - & - 
  & 9.59 & 4.77 
  & 4.85 & 2.11 
  & - & - 
  & 2.26 & 1.14 
  & - & - 
  & 7.42 & 3.58 \\

C & \textbf{0.61} & \textbf{0.35} 
  & - & - 
  & 2.18 & 1.07 
  & 2.07 & 0.98 
  & - & - 
  & 1.93 & 0.76 
  & - & - 
  & 1.86 & 0.92 \\

D & \textbf{0.90} & \textbf{0.44} 
  & - & - 
  & 2.06 & 0.85 
  & 12.48 & 5.60 
  & - & - 
  & 2.74 & 1.17 
  & - & - 
  & 1.65 & 0.64 \\

E & \textbf{1.14} & \textbf{0.75} 
  & - & - 
  & 3.97 & 1.70 
  & 6.44 & 3.00 
  & - & - 
  & 7.38 & 3.10 
  & - & - 
  & 3.40 & 1.49 \\

\midrule

Avg. & \textbf{0.66} & \textbf{0.38} 
     & - & - 
     & 3.81 & 1.82 
     & 5.81 & 2.62 
     & - & - 
     & 2.89 & 1.25 
     & - & - 
     & 3.13 & 1.44 \\

\midrule 
\multicolumn{17}{l}{\footnotesize *Variant with the loop closure module activated.} \\
\bottomrule
\end{tabular}
\end{table*}

\begin{table}[htbp]
\centering
\footnotesize
\setlength{\tabcolsep}{2.8pt}
\caption{Trajectory re-entry position errors across GNSS-degraded intervals (Unit: m).}
\label{tab:gap_closure}
\begin{tabular}{lccccc}
\toprule
Interval & Inland-LOAM & KISS-ICP & HDL-SLAM & LeGO-LOAM & LOAM \\
\midrule
$A_{I-1}$ & \textbf{0.15} & 0.26 & 1.03 & 0.17 & 0.17 \\
$A_{I-2}$ & \textbf{0.09} & 0.55 & 0.53 & 0.11 & 0.45 \\
$A_{I-3}$ & \textbf{0.09} & 0.17 & 1.68 & 0.29 & 0.22 \\
$C_{I-1}$ & 0.61 & 2.39 & 2.42 & 1.42 & \textbf{0.23} \\
\bottomrule
\end{tabular}
\end{table}

\subsection{Ablation Study}
\label{sec4_ablation}

Before analyzing the ablation results, we first verified the water plane observations used in the water surface constraint. Valid water surface planes were estimated by RANSAC in all frames of the five test areas, with median inlier ratios of 0.73--0.84 and median fitting RMSE values of approximately 2.47--3.03 cm. These results show that the proposed extraction method consistently captures the water plane geometry in the evaluated inland waterway scenes.

To evaluate the contribution of the main components in Inland-LOAM, we conducted an ablation study with four configurations: a baseline configuration based on LeGO-LOAM, the baseline with the proposed feature extraction, the configuration with the water surface constraint, and the full Inland-LOAM configuration.

\begin{table}[!t]
\centering
\footnotesize
\setlength{\tabcolsep}{3pt}
\caption{Component ablation study of trajectory accuracy (Unit: m).}
\label{tab:ablation}

\begin{tabular}{lcccccccc}
\toprule
& \multicolumn{2}{c}{Baseline}
& \multicolumn{2}{c}{+ Features}
& \multicolumn{2}{c}{+ Water}
& \multicolumn{2}{c}{Full} \\
\cmidrule(lr){2-3}
\cmidrule(lr){4-5}
\cmidrule(lr){6-7}
\cmidrule(lr){8-9}
Area
& RMSE & STD
& RMSE & STD
& RMSE & STD
& RMSE & STD \\
\midrule
A & 0.14 & 0.08 & 0.11 & 0.06 & \textbf{0.09} & \textbf{0.05} & \textbf{0.09} & \textbf{0.05} \\
B & 2.26 & 1.14 & 2.13 & 1.09 & 0.62 & 0.31 & \textbf{0.58} & \textbf{0.30} \\
C & 1.93 & 0.76 & 2.05 & 1.12 & \textbf{0.61} & \textbf{0.35} & \textbf{0.61} & \textbf{0.35} \\
D & 2.74 & 1.17 & 0.98 & \textbf{0.39} & 0.96 & 0.44 & \textbf{0.90} & 0.44 \\
E & 7.38 & 3.10 & 2.18 & 0.97 & 1.22 & 0.81 & \textbf{1.14} & \textbf{0.75} \\
\midrule
Avg. & 2.89 & 1.25 & 1.49 & 0.73 & 0.70 & 0.39 & \textbf{0.66} & \textbf{0.38} \\
\bottomrule
\end{tabular}

\vspace{1mm}
\parbox{\columnwidth}{\footnotesize
Baseline: configuration based on LeGO-LOAM; Features: proposed feature extraction; Water: water surface constraint; Full: proposed feature extraction, water surface constraint, and Huber robust noise model with $\delta=1.345$.
}
\end{table}

Table~\ref{tab:ablation} reports the incremental effect of the proposed components. The feature extraction strategy reduces the average RMSE from 2.89 m to 1.49 m, with the largest gains in Areas D and E, where the added rough and planar features are useful for vegetation and mixed shoreline structures. In Area C, the RMSE slightly increases from 1.93 to 2.05 m, possibly because regular man-made structures already provide sufficient constraints for the baseline and some additional features become redundant for odometry. However, the proposed feature extraction reduces the RMSE in the other four test areas. The extracted rough and planar features are also used in the structural semantic mapping stage. Whether adaptive feature selection can reduce this redundancy while maintaining sufficient geometric constraints remains to be investigated.

Adding the water surface constraint further reduces the average RMSE to 0.70 m, confirming that the water surface provides a global reference for reducing drift in weakly structured waterway scenes. The full configuration achieves the lowest average RMSE and STD. The gain from the Huber robust noise model is modest, since the water surface planes are already fitted reliably in most frames and only a small number of large residuals need to be downweighted. Nevertheless, the Huber robust noise model further improves the results in Areas B, D, and E, where vegetation, mixed shoreline geometry, and longer weakly constrained sections make the trajectory more sensitive to water plane observations.

\subsection{Structural Semantic Mapping Evaluation}
\label{sec4_3}

\begin{figure*}[!t]
  \centering
  \includegraphics[width=0.85\textwidth]{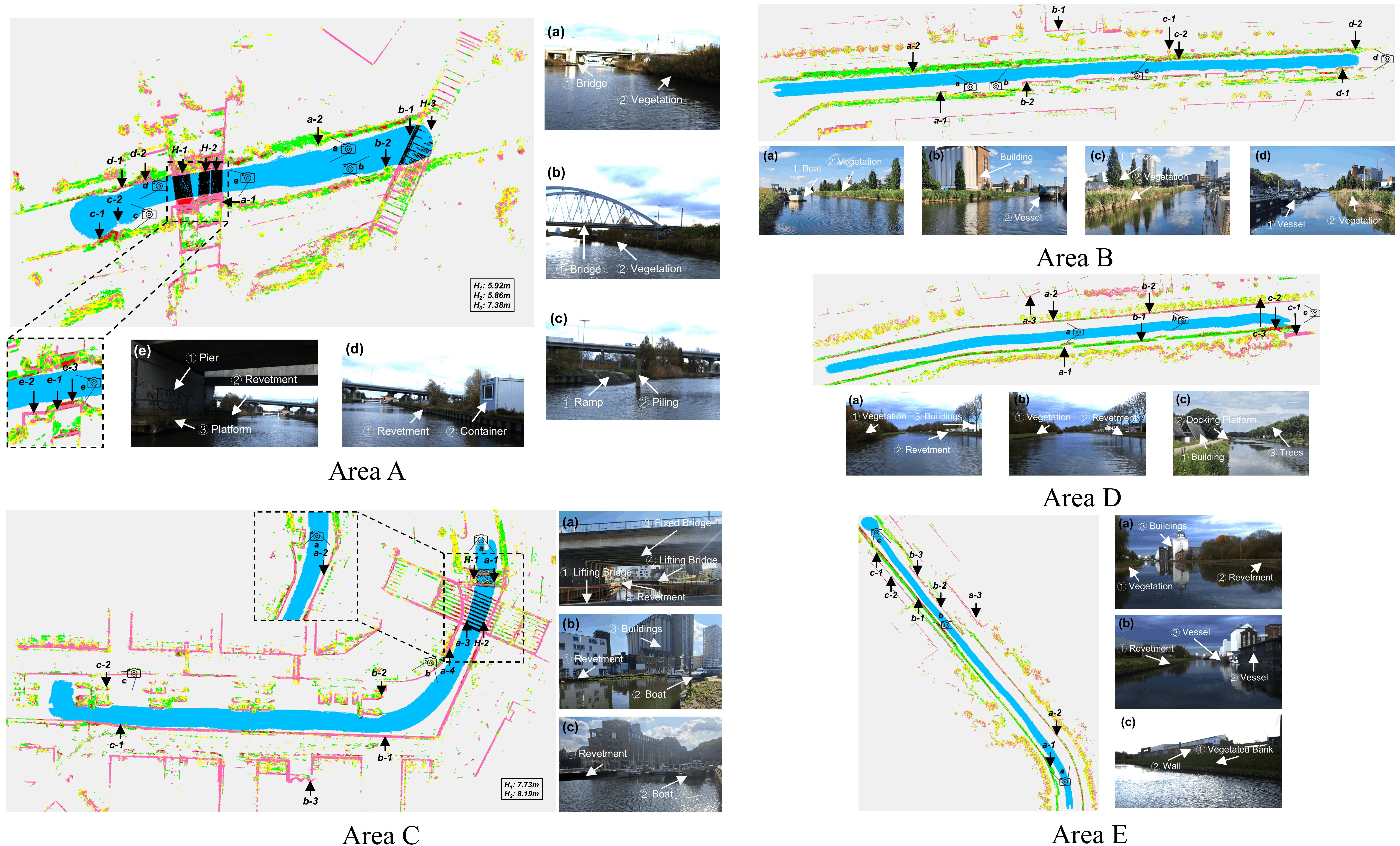}
  \caption{Structural semantic maps generated by Inland-LOAM in Areas A--E. Each scenario is presented with corresponding real-world photographs from multiple perspectives, annotated to explain key features. The color encoding follows the structural semantic classes defined in Section~\ref{subsec3_4_2}: red for HP, pink for VP, green for CH, yellow for CV, blue for OW, and black for UB. For Areas A and C, detailed semantic maps of bridge areas are included, with the identified clearance height explicitly labeled H.}
  \label{fig:semantic_map}
\end{figure*}

\subsubsection{Qualitative Analysis of Structural Semantic Maps}

Figure~\ref{fig:semantic_map} provides a qualitative overview of the structural semantic maps in Areas A--E. The results demonstrate that the mapping system successfully preserves critical navigation-related structures while capturing diverse local shoreline characteristics. In Area A, which contains two fixed bridges, OW follows the narrow canal, and UB is detected at the bridge passages with clearance estimates H-1, 2, and 3. The same area also shows several typical shoreline objects: boat ramps are classified as HP at c-1, shore platforms as HP at e-3, containers as VP at d-2, bridge piers as VP at e-1, vegetated banks as CH/CV at a-2, and piling structures as CV at c-2. Similar distinctions appear in Area E, where low vegetated banks and higher revetment structures are separated at a-1 and a-2, while vessels at b-2 and b-3 and walls at c-2 are represented as VP.

Areas B and D mainly demonstrate the behavior of the geometric rules in environments dominated by vegetation. In Area B, the large building at b-1 is classified as VP, whereas most shoreline regions covered with reeds are assigned to CH or CV. The occasional small HP and VP regions within these vegetated areas, for example near c-2 and d-2, likely result from LiDAR returns from exposed ground, sparse vegetation, or locally planar parts of dense vegetation. In Area D, the map separates a building overhang (VP) and an adjacent docking platform (HP) at c-1 and c-2, and distinguishes two
different shoreline structures at b-1 and b-2, classified as
CH/CV and VP, respectively. The former mainly corresponds
to low vegetation along the bank, whereas the latter appears
as a higher and more continuous stone revetment.

Area C represents the urban port environment. VP is clearly produced on building facades at b-3, vessels at b-2 and c-2, and revetments at b-1 and c-1, while UB identifies the navigable water beneath the bridge structures. Around the lifting bridge at H-1, local color mixing appears because the bridge moved during data collection, so different scans observed different structural states. Overall, the semantic map could be interpreted as a navigation-oriented structural layer. For path planning, OW provides the navigable corridor; UB marks regions requiring height awareness; HP/VP can support docking, boundary detection, and localization; and CH/CV indicate rough or vegetated banks that should be avoided or treated with a higher traversal cost.

\subsubsection{Quantitative Evaluation of Structural Semantic Classes}

\begin{figure}[!t]
  \centering
  \includegraphics[width=\columnwidth]{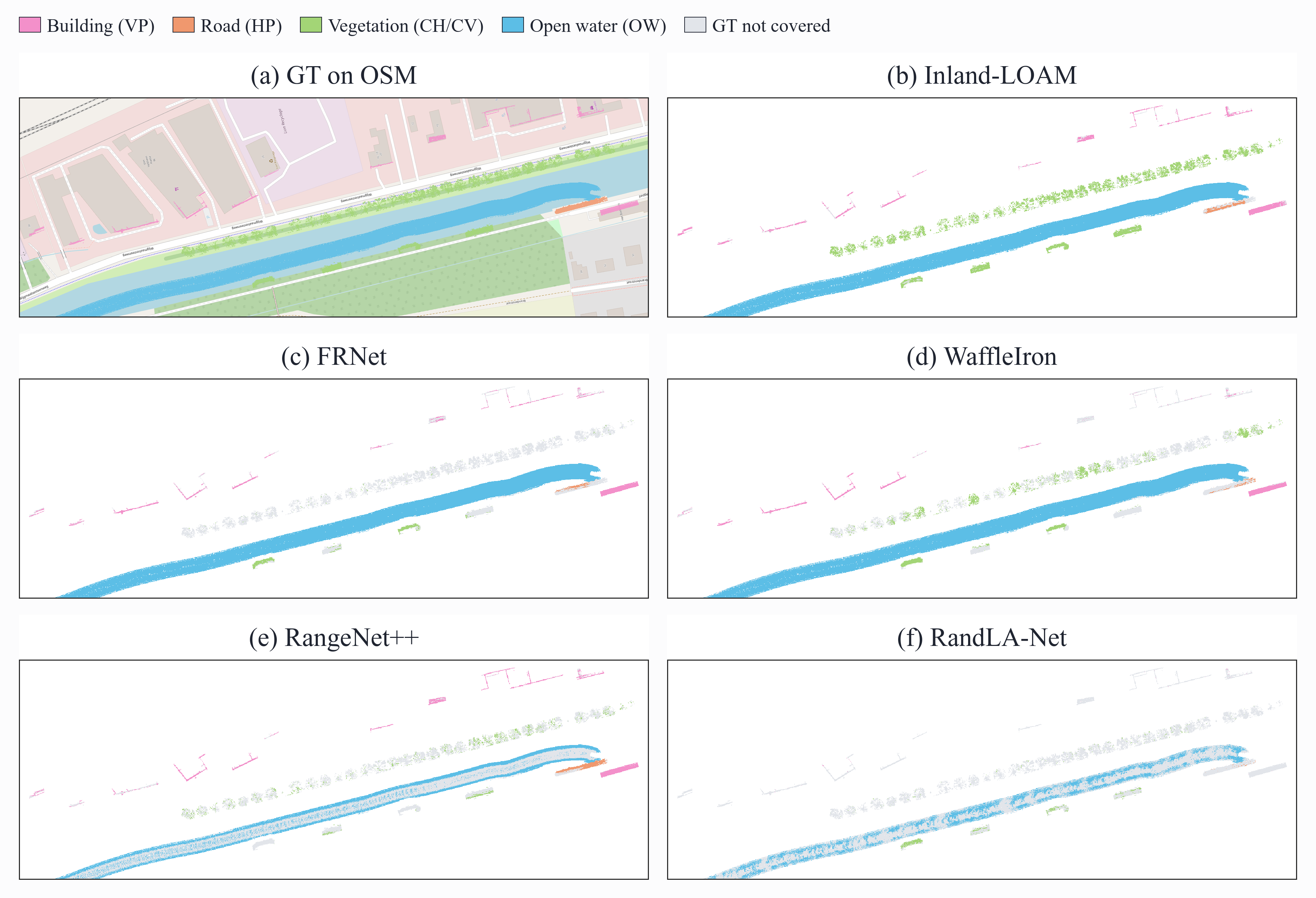}
  \caption{Comparison of the annotated ground truth (GT) masks and the projected predictions of all evaluated methods in Area D.}
  \label{fig:area_d_roi_projection}
\end{figure}

Following the qualitative analysis, we further evaluate the structural semantic classes in Area D. This area contains open water, densely vegetated banks, trees, docking platforms, and vertical structures, providing a comprehensive testbed for evaluating semantic classification. Lacking dense point-level ground truth, we manually annotated a Region of Interest (ROI) using OpenStreetMap (OSM)\footnote{\url{https://www.openstreetmap.org}} and Google Earth as references. The resulting annotated ground truth (GT) masks and the projected predictions of all evaluated methods are shown in Figure~\ref{fig:area_d_roi_projection}. To evaluate semantic performance without the influence of odometry drift, all evaluated baselines are tested on the same point cloud map generated by the Inland-LOAM trajectory. 

We conduct a domain transfer evaluation with four learning methods: FRNet~\cite{xu2025frnet}, WaffleIron~\cite{puy2023using}, RangeNet++~\cite{milioto2019rangenet++}, and RandLA-Net~\cite{hu2020randla}. All four methods use public weights pretrained on SemanticKITTI without further training on inland waterway data. To compare them within the same ROI framework, we map the closest SemanticKITTI labels to our structural classes: \textit{building} and \textit{fence} are mapped to VP; \textit{road}, \textit{other-ground}, and \textit{terrain} are mapped to HP; and \textit{vegetation} is mapped to CH/CV. Because SemanticKITTI has no water class, OW is evaluated in a separate binary ROI comparison in which \textit{road} is used as the closest available surface proxy. Thus, the \textit{road} predictions used in the HP and OW evaluations are compared independently with the corresponding reference masks and do not constitute a single multiclass label mapping. This mapping is used only for ROI coverage comparison, not as a strict semantic equivalence. All point predictions are then projected into the same 2D grid and compared with the annotated reference masks.

\begin{table}[!t]
\centering
\footnotesize
\setlength{\tabcolsep}{3.0pt}
\renewcommand{\arraystretch}{1.05}
\caption{ROI IoU comparison in the annotated regions (Domain transfer evaluation).}
\label{tab:seg4_multiclass_roi_iou}
\begin{tabular}{lccccc}
\toprule
Class &
\makecell{Inland\\LOAM} &
FRNet &
\makecell{Waffle\\Iron} &
\makecell{Range\\Net++} &
\makecell{RandLA\\Net} \\
\midrule
Building (VP) & \textbf{0.794} & 0.676 & 0.580 & 0.779 & 0.009 \\
Road (HP) & 0.539 & 0.262 & 0.203 & \textbf{0.695} & 0.056 \\
Vegetation (CH/CV) & \textbf{0.902} & 0.080 & 0.321 & 0.166 & 0.081 \\
Open Water (OW) & \textbf{1.000} & 0.990 & 0.978 & 0.444 & 0.350 \\
\bottomrule
\end{tabular}
\end{table}

Table~\ref{tab:seg4_multiclass_roi_iou} shows that Inland-LOAM achieves the highest IoU scores for the VP, CH/CV, and OW classes, while RangeNet++ obtains the best result for HP. The lower HP score reveals a limitation of the rule-based conversion: its reliance on local geometry at the expense of global context. As a result, local elements such as weeds or uneven steps on a docking platform can easily disrupt the planar criteria, preventing the complete extraction of the horizontal surface. Despite this, the results across VP, CH/CV, and OW confirm that our method extracts the main structural regions. Overall, these results illustrate the challenges of directly transferring the evaluated models trained on road scenes to inland waterways. The proposed approach provides structural semantic labels without requiring pointwise semantic annotations or domain retraining.

\begin{table}[!t]
\centering
\footnotesize
\setlength{\tabcolsep}{2.0pt}
\caption{Structural decision rules used in structural semantic map conversion.}
\label{tab:ssm_rule_params}
\begin{tabular}{lccc}
\toprule
Rule & Count ($N$) & Plane ratio ($R$) & Slope ($m$; $\theta$) \\
\midrule
A (VP) & $>9$  & $\ge 0.50$ & $>0.45$ ($>56.3^\circ$) \\
B (VP) & $>10$ & $\ge 0.65$ & $>0.20$ ($>33.7^\circ$) \\
C (HP) & $>10$ & $\ge 0.80$ & $<0.15$ ($<26.6^\circ$) \\
D (HP) & $>7$  & $\ge 0.75$ & $\le 0.10$ ($\le 18.4^\circ$) \\
E (CV) & --    & --         & $>0.15$ ($>26.6^\circ$) \\
F (CH) & --    & --         & otherwise \\
\bottomrule
\end{tabular}

\vspace{1mm}
\parbox{\columnwidth}{\footnotesize The slope angle is computed as $\theta=\tan^{-1}(m/r_g)$ using the 2D grid resolution $r_g=0.30\,\mathrm{m}$.}
\end{table}

\subsubsection{Sensitivity Analysis of Structural Semantic Rules}

Finally, to verify that the structural semantic conversion is not overly dependent on specific threshold choices, we performed a local sensitivity analysis using the same annotated ROI and map input as in the quantitative evaluation. The structural decision rules and their default parameters used in the map conversion are summarized in Table~\ref{tab:ssm_rule_params}. Rules A and B generate VP, Rules C and D generate HP, and the remaining observed non-water cells are assigned to CH and CV. OW and UB are treated as fixed overrides and are not perturbed in this test. This test is not intended to tune the parameters in Area D. Instead, it evaluates the robustness of the default thresholds and shows how perturbing one rule changes the final class assignment.

\begin{figure}[!t]
\centering
  \includegraphics[width=\columnwidth]{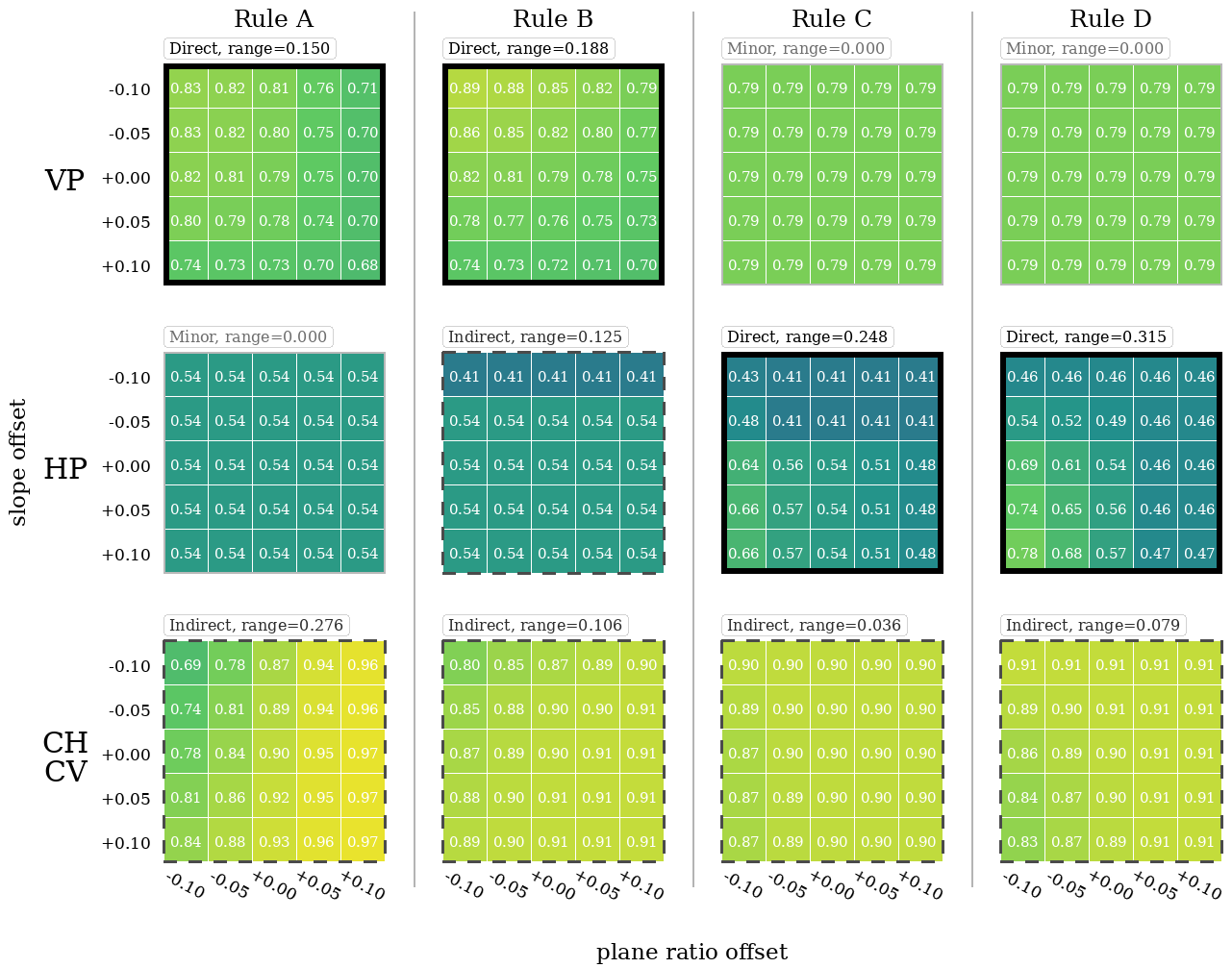}
  \caption{Sensitivity of structural decision rules in the annotated ROI. Each heatmap reports ROI IoU under plane ratio and slope threshold offsets from the default setting.}
  \label{fig:vp_branch_sensitivity}
\end{figure}

Figure~\ref{fig:vp_branch_sensitivity} shows that the algorithm is generally stable under threshold changes. Rules A and B mainly affect VP: moderately relaxing their thresholds improves the VP IoU from the default $0.79$ to peaks of $0.83$ and $0.89$. Rules C and D primarily affect the HP class. They cause wider IoU fluctuations ($0.41$--$0.66$ and $0.46$--$0.78$) because docking platforms, flat surfaces, and low bank regions can share similar local planar geometry. Ultimately, while fine-tuning thresholds can improve a single class locally, it may also shift errors to neighboring classes.

\subsection{Shoreline Boundaries Comparison with Reference Maps}
\label{sec4_4}

\begin{figure}[!t]
  \centering
  \includegraphics[width=0.98\columnwidth]{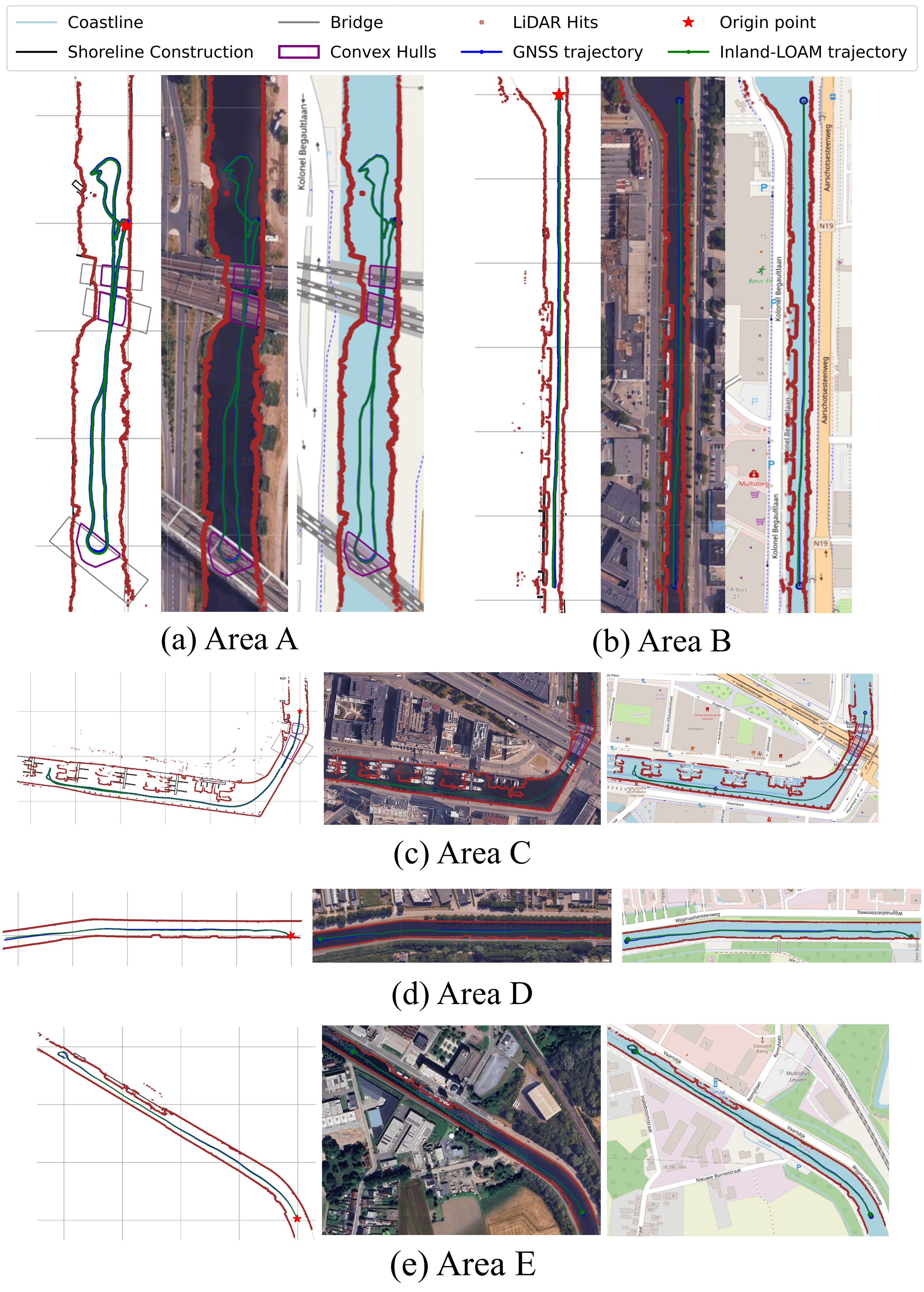}
  \caption{The shoreline boundaries generated by Inland-LOAM were projected onto the IENC, Google Earth, and OpenStreetMap in Areas A--E.}
  \label{fig:google_map}
\end{figure}

To evaluate the effectiveness of the boundaries derived from structural semantic mapping, we performed a comparative analysis with three widely used reference maps: Google Earth, OSM, and IENCs. As illustrated in Figure~\ref{fig:google_map}, each reference map was created using data from a different period: the Google Earth imagery was dated June 16, 2023 or earlier; OSM is based on a continuously updated database; and the IENC data were last modified on July 1, 2021.

Across Areas A--E, the extracted shoreline boundaries generally follow stable physical structures such as revetments, quay walls, canal banks, and moored vessels. 
In Area A, the boundaries (red hits) align well with artificial shoreline structures in both IENC and Google Earth, while local differences near vegetated banks are mainly caused by seasonal reed growth or minor erosion not represented in the static maps. In Area B, Inland-LOAM captures the canal boundary and long-term moored vessels along the waterway. Area C shows a more dynamic port case: although the canal boundary aligns with the reference maps, the detected vessel layout differs from older map sources and is closer to the updated OSM, indicating the potential of Inland-LOAM for maintaining up-to-date navigation charts. The additional Areas D and E further show that the boundary extraction remains consistent in vegetation-dominated and mixed shoreline environments.

\subsection{Runtime Analysis}
\label{sec4_5}

The LOAM module runs in real time. The front-end, including preprocessing, plane fitting, feature extraction, and LiDAR odometry, requires 18.3 ms in total, with 7.8 ms for preprocessing and plane fitting and 10.5 ms for feature extraction and odometry. The back-end mapping module requires 37.2 ms. These values are below the acquisition interval of the 10 Hz LiDAR and the 2 Hz map generation interval, respectively.

\begin{figure}[!t]
\centering
  \includegraphics[width=0.85\columnwidth]{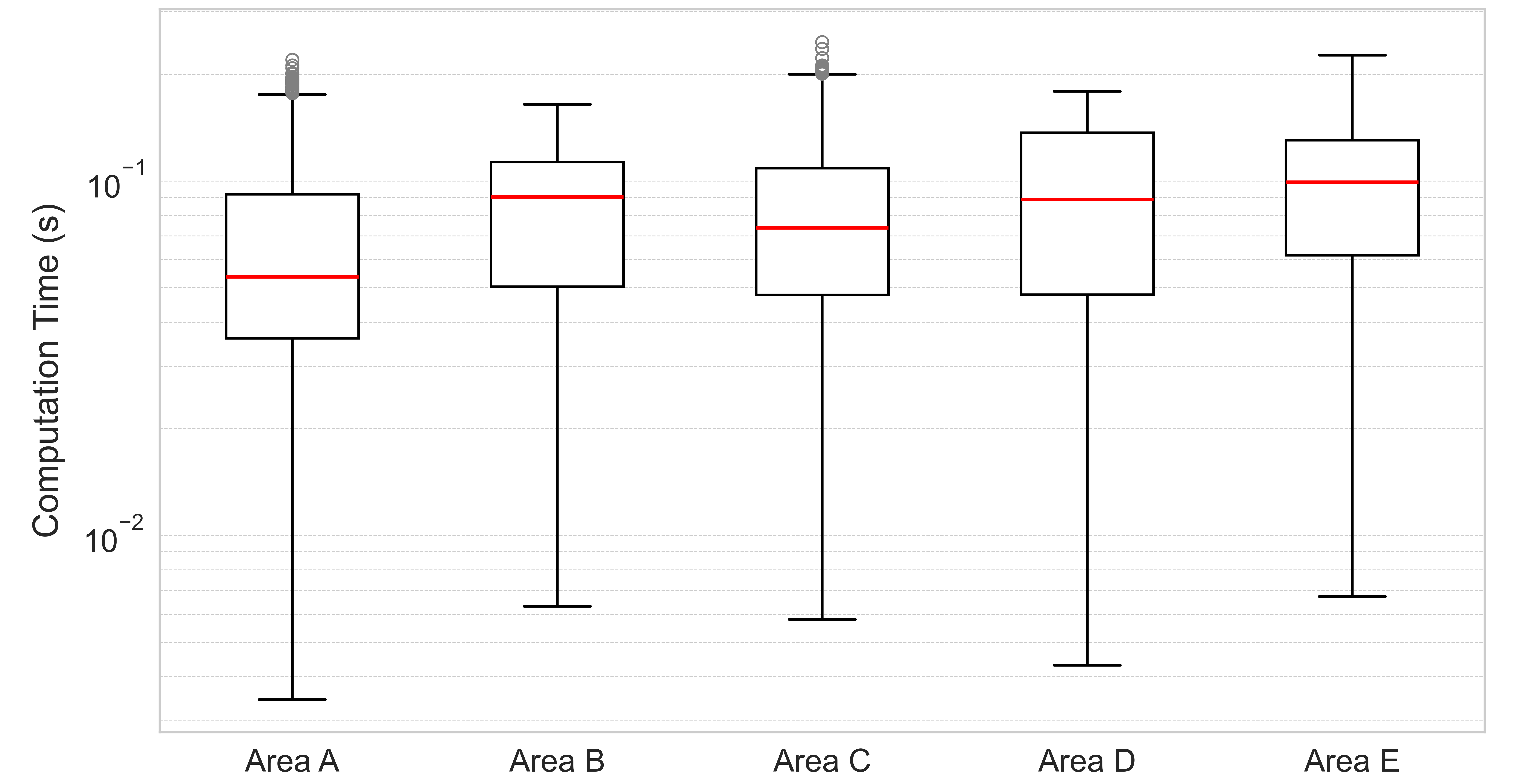}
  \caption{The computation time to update the 2D map generated from the voxel map under different exploration areas.}
  \label{fig:conversion_time}
\end{figure}

Figure~\ref{fig:conversion_time} reports the runtime of converting the voxel map into the 2D structural semantic map in Areas A--E. The conversion time mainly depends on the number of voxel map regions updated in each callback. Area A shows the lowest runtime because its loop-back trajectory revisits previously mapped segments, whereas Areas B--E involve broader open-area exploration and therefore require larger map updates. Nevertheless, the average conversion time remains below 0.1 s across all five areas, indicating that the conversion module can update the structural semantic map online within the 2 Hz map generation interval. This provides a sufficient computational margin for the tested USV operation.

The IENC integration module is executed offline because it is intended for long-term map maintenance and sharing. Therefore, it does not introduce additional computational cost during online navigation.

\section{Conclusion}
\label{sec5}
In this paper, we propose Inland-LOAM, an integrated LiDAR-only SLAM framework specifically designed for robust odometry, mapping, and structural semantic map generation in challenging inland waterway environments. Addressing the limitations of conventional SLAM methods, our contributions include an enhanced feature extraction approach and a joint optimization method that uses the water surface as a globally consistent planar reference to reduce drift. Furthermore, we detail an end-to-end pipeline that transforms dense 3D point clouds into 2D structural semantic maps. The proposed method, which combines voxel-based semantic segmentation with subsequent 2D mapping, enables real-time estimation of critical navigational parameters, such as bridge clearances, and provides delineated shorelines and spatial attributes of bridges that can support integration with standard navigational charts and chart update workflows.

Experiments demonstrate that Inland-LOAM improves localization accuracy across the evaluated routes and generates structural semantic maps that capture the main waterway features. The resulting semantic maps and georeferenced overlays can provide useful inputs for situational awareness and downstream navigation functions.

However, our method has certain limitations. Purely geometry-based classification can be ambiguous in regions with complex natural textures, such as densely vegetated areas, leading to occasional local misclassifications. Additionally, the current static mapping approach faces challenges when encountering dynamic elements, as evidenced by the aliasing artifacts caused by moving lifting bridges. Reliable detection of the water surface itself can also be compromised in highly dynamic circumstances, such as in the presence of nearby moving vessels, which can degrade the effectiveness of our global planar constraint and introduce additional drift. This highlights potential vulnerabilities when operating in environments with significant dynamic infrastructure or dense vessel traffic. In very wide waterways or sections with low and distant banks, stable shoreline features may become sparse. Although the water surface constraint provides a useful vertical reference, it does not constrain lateral motion or heading drift, which still depend on sufficient shoreline geometry. 

Future research will focus on addressing these limitations. For adaptive feature selection, we will investigate how local feature density and geometric diversity can guide the number and spatial distribution of selected points while preserving sufficient geometric constraints. For loop closure, we will investigate verification criteria that consider scan overlap, the spatial distribution of geometric correspondences, and uncertainty in the estimated relative pose before adding loop constraints to the pose graph. To further improve semantic classification accuracy, we will explore the integration of additional sensor modalities, particularly RGB cameras. Color and texture information could help resolve semantic ambiguities among geometrically similar but semantically distinct surfaces. To improve robustness in dynamic environments, future work will investigate dynamic object segmentation, temporal occupancy decay, consistency checks across sessions, object motion estimation for vessels and moving infrastructure, and fusion with camera or radar. Dynamic navigation layers should also be kept separate from long-term chart maintenance layers, so that moving objects can support collision avoidance without corrupting the static geospatial overlay. Moreover, the generated semantic maps and geospatial overlays can support higher-level autonomous navigation functions, such as bridge clearance checking, docking assistance, chart validation, semantic-guided localization using stable landmarks, and path planning based on semantic costmaps.

\section*{Acknowledgment}
The authors would like to thank the Editor-in-Chief, an Associate Editor, and anonymous referees for their invaluable comments and suggestions.
 
\bibliographystyle{IEEEtran}
\bibliography{ref}

\begin{IEEEbiography}[{\includegraphics[width=1in,height=1.25in,clip,keepaspectratio]{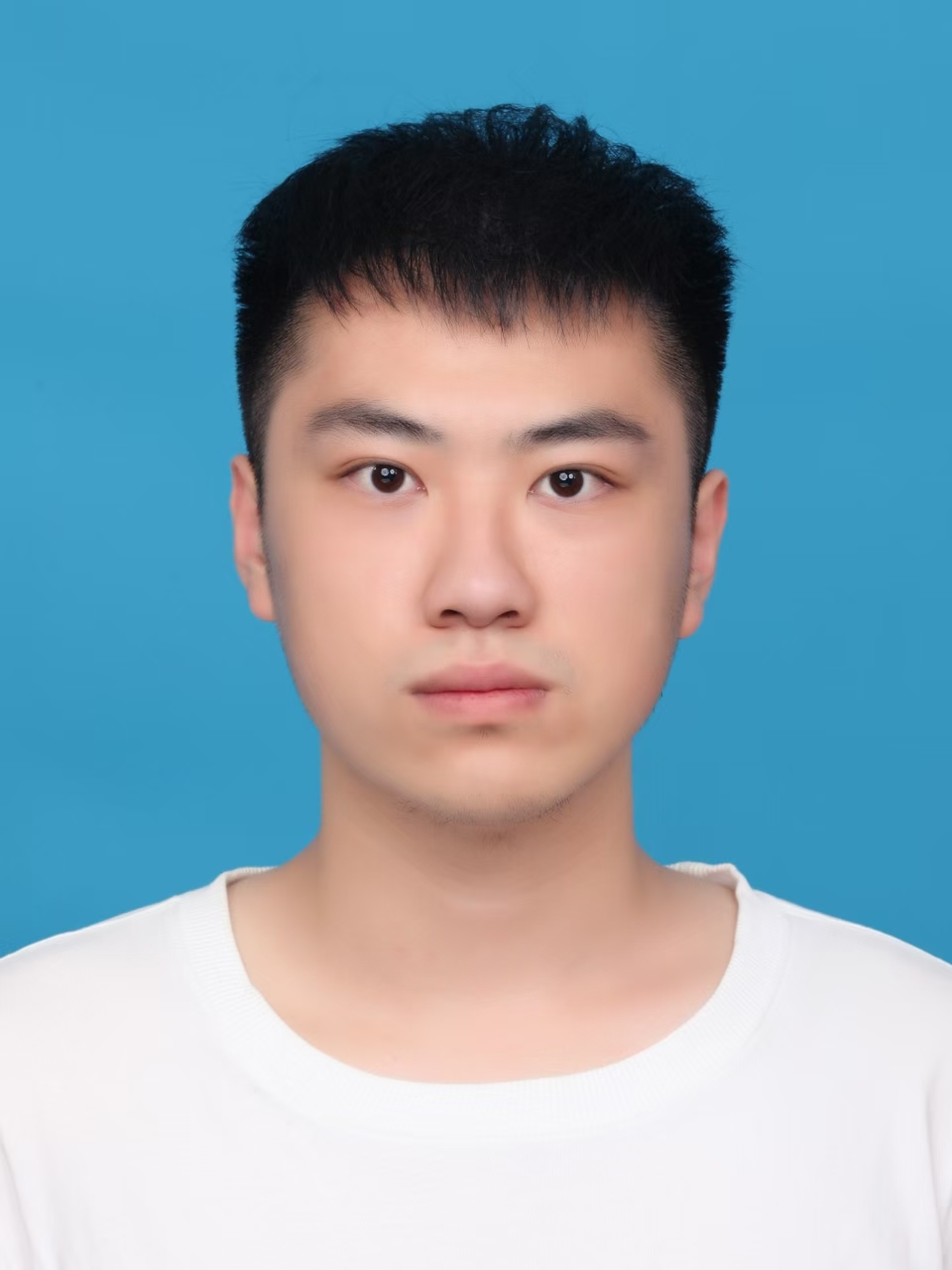}}]{Zhongbi Luo}
received the B.E. degree in mechanical design, manufacturing and automation from Huazhong Agricultural University, Wuhan, China, in 2020, and the M.Sc. degree in mechanical engineering design from The University of Manchester, Manchester, U.K., in 2022. He is currently working toward the Ph.D. degree with the Division of Robotics, Automation and Mechatronics, Department of Mechanical Engineering, KU Leuven, Leuven, Belgium. He is an Early-Stage Researcher with the Marie Sk\l{}odowska-Curie ITN AUTOBarge project. His research interests include scene understanding, simultaneous localization and mapping, and intelligent navigation for autonomous vessels.
\end{IEEEbiography}

\begin{IEEEbiography}[{\includegraphics[width=1in,height=1.25in,clip,keepaspectratio]{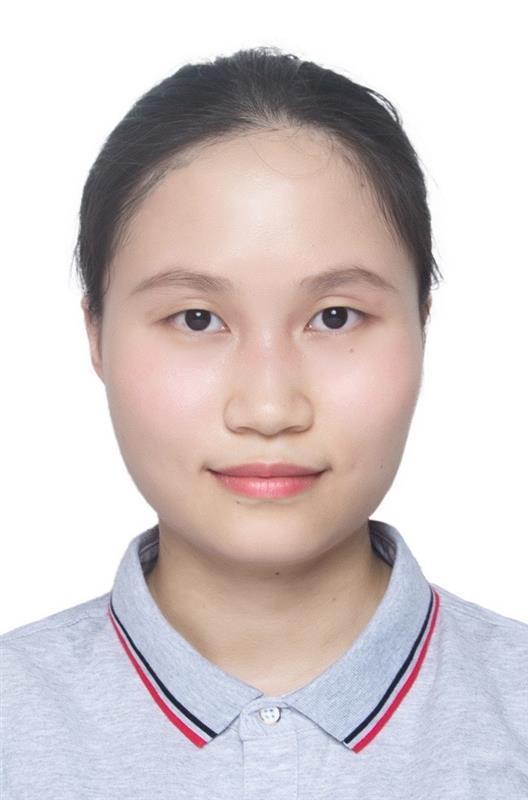}}]{Yunjia Wang}
received a B.E. degree in information engineering and an M.S. degree in information and communications engineering, both from Zhejiang University, Hangzhou, China. After two years of industrial experience as an applied AI scientist at Bosch Center for Artificial Intelligence, she is currently a PhD researcher of the M-Group and the Declarative Languages and Artificial Intelligence (DTAI) research group at the KU Leuven Bruges campus in Belgium, within the Department of Computer Science. She is also affiliated with Flanders Make, the strategic research center for the manufacturing industry. Her current research interests include safety assurance and machine learning model robustness for autonomous vessels.
\end{IEEEbiography}

\begin{IEEEbiography}[{\includegraphics[width=1in,height=1.25in,clip,keepaspectratio]{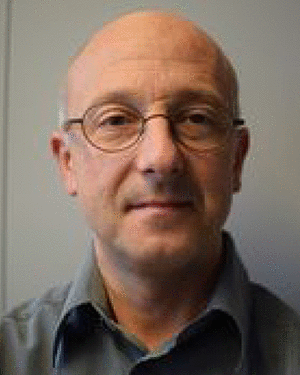}}]{Jan Swevers}
received the M.S. degree in electrical engineering and the Ph.D. degree in mechanical engineering from KU Leuven, Leuven, Belgium, in 1986 and 1992, respectively. He is currently a Full Professor with the Department of Mechanical Engineering, KU Leuven. His research interests include real-time/embedded optimization based control and iterative learning control of mechatronic systems including robotic manipulators, advanced friction modeling and compensation, optimal experiment design and identification of nonlinear mechatronic systems, and robust feedback control design for linear time-invariant and parameter-varying systems.
\end{IEEEbiography}

\begin{IEEEbiography}[{\includegraphics[width=1in,height=1.25in,clip,keepaspectratio]{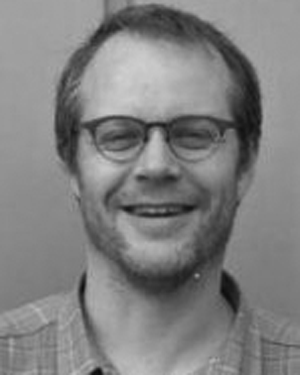}}]{Peter Slaets}
received the MSc degree in electro\-tech\-ni\-cal-\hspace{0pt}me\-chan\-i\-cal engineering, specialization in data mining and automation, in 2002 from KU Leuven, Leuven, Belgium, and the PhD degree in applied sciences from KU Leuven, in 2008 with the title: `Geometric 3D Model Building from Sensor Measurements Collected during Compliant Motion: Stochastic Filtering and Hardware Architectures'. In 2005, he became a lecturer with the Katholieke Hogeschool Limburg (KHLIM), Diepenbeek, Belgium, and the Katholieke Hogeschool Kempen (KHK), Geel, Belgium, where he teaches courses in digital electronics, control, and automation. He is currently an associate professor in the intelligent mobile platform research group with the Department of Mechanical Engineering, KU Leuven. His research focuses on modeling, Bayesian estimation techniques, and mobile platforms in general, with applications in autonomous inland shipping and health monitoring in particular.
\end{IEEEbiography}

\begin{IEEEbiography}[{\includegraphics[width=1in,height=1.25in,clip,keepaspectratio]{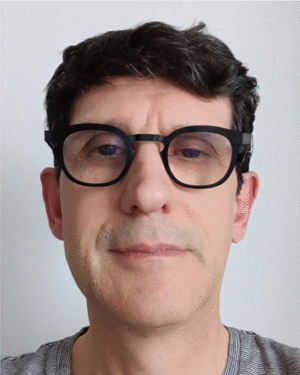}}]{Herman Bruyninckx}
received the master's degrees in mathematics, computer science, and mechatronics, in 1984, 1987, and 1988, respectively, and the Ph.D. degree in mechanical engineering, in 1995, from KU Leuven, Leuven, Belgium. He is currently a full-time Full Professor with KU Leuven, and part time with the Eindhoven University of Technology, Eindhoven, The Netherlands. His research focus in both places is on the composability of the most advanced, knowledge driven algorithms for the dynamics of motion control of complex robotics applications, with distributed sensor processing and resource monitoring. The complementary objectives are to realise such systems with the least amount of resource costs, with ``good enough'' quality, and with full explainability. He held visiting research positions with the University of Pennsylvania, Philadelphia, Stanford University, CA, USA, and the Kungl Tekniska Hogskolan, Stockholm, Sweden. From 2007 to 2015, he was leading the European academic community, first as Coordinator of the European Robotics Research Network EURON, and later as Vice-President Research of euRobotics. He was the Principal Investigator in about two dozen European research projects on robotics, with a focus in the recent years on the systems and software engineering aspects of multiagent robotic skills.
\end{IEEEbiography}

\end{document}